%% file: main.tex
\documentclass[10pt,twocolumn,letterpaper]{article}

\usepackage[pagenumbers]{iccv} 


\usepackage[accsupp]{axessibility}
\usepackage{booktabs}
\usepackage{algorithm}
\usepackage{algpseudocode}
\usepackage{multirow}
\usepackage{array}
\usepackage{float}
\usepackage{bbm}
\usepackage{tablefootnote}
\usepackage{comment}
\newcommand{\drule}{\specialrule{0.2pt}{1pt}{1pt}%
            \specialrule{0.2pt}{0pt}{\belowrulesep}%
            }
  {\begin{list}{}%
          {\setlength{\leftmargin}{#1}}%
          \item[]%
  }
  {\end{list}}

\usepackage{pifont}
\usepackage{xcolor}
\usepackage{pdfrender}
\newcommand{\xmark}{\ding{55}}%
\usepackage{colortbl}
  
\newcommand*{\affmark}[1][*]{\textsuperscript{#1}}

\usepackage{kotex}

\definecolor{applegreen}{rgb}{0.55, 0.71, 0.0}

\definecolor{forestgreen}{rgb}{0.13, 0.55, 0.13}

\newcommand\blfootnote[1]{%
  \begingroup
  \renewcommand\thefootnote{}\footnote{#1}%
  \addtocounter{footnote}{-1}%
  \endgroup
}

\definecolor{iccvblue}{rgb}{0.21,0.49,0.74}
\usepackage[pagebackref,breaklinks,colorlinks,allcolors=iccvblue]{hyperref}

\def\confName{ICCV}
\def\confYear{2025}

\title{Details Matter for Indoor Open-vocabulary 3D Instance Segmentation\vspace*{-0.3cm}}

\author{
Sanghun Jung\affmark[1,2]$^\dagger$ \; Jingjing Zheng\affmark[2] \; Ke Zhang\affmark[2] \; Nan Qiao\affmark[2] \; Albert Y. C. Chen\affmark[2] \; Lu Xia\affmark[2] \; Chi Liu\affmark[2] \\ Yuyin Sun\affmark[2] \; Xiao Zeng\affmark[2] \; Hsiang-Wei Huang\affmark[1] \; Byron Boots\affmark[1] \; Min Sun\affmark[2,3] \; Cheng-Hao Kuo\affmark[2] \vspace{0.2cm}\\
\affmark[1]University of Washington \; \affmark[2]Amazon Lab126 \; \affmark[3]National Tsing Hua University\\
}

\begin{document}
\maketitle
\input{sec/0_abstract}    
\blfootnote{$\dagger$ The work is done during the internship at Amazon Lab126.\vspace*{-0.8cm}}
\input{sec/1_intro}
\input{sec/2_related_work}
\input{sec/3_method}

\input{sec/4_experiments}

\input{sec/5_conclusion}
\clearpage
{
    \small
    \bibliographystyle{ieeenat_fullname}
    \bibliography{main}
}

\clearpage
\appendix
\input{sec/X_suppl}

\end{document}

%% file: sec/0_abstract.tex
\begin{abstract}
Unlike closed-vocabulary 3D instance segmentation that is often trained end-to-end, open-vocabulary 3D instance segmentation (OV-3DIS) often leverages vision-language models (VLMs) to generate 3D instance proposals and classify them.
While various concepts have been proposed from existing research, we observe that these individual concepts are not mutually exclusive but complementary.
In this paper, we propose a new state-of-the-art solution for OV-3DIS by carefully designing a recipe to combine the concepts together and refining them to address key challenges.
Our solution follows the two-stage scheme: 3D proposal generation and instance classification.
We employ robust 3D tracking-based proposal aggregation to generate 3D proposals and remove overlapped or partial proposals by iterative merging/removal.
For the classification stage, we replace the standard CLIP model with Alpha-CLIP, which incorporates object masks as an alpha channel to reduce background noise and obtain object-centric representation.
Additionally, we introduce the standardized maximum similarity (SMS) score to normalize text-to-proposal similarity, effectively filtering out false positives and boosting precision.
Our framework achieves state-of-the-art performance on ScanNet200 and S3DIS across all AP and AR metrics, even surpassing an end-to-end closed-vocabulary method.
\end{abstract}

%% file: sec/1_intro.tex
\vspace{-0.5cm}
\section{Introduction}
\label{sec:intro}
\begin{figure}[t]
  \centering
  \includegraphics[width=0.8\linewidth]{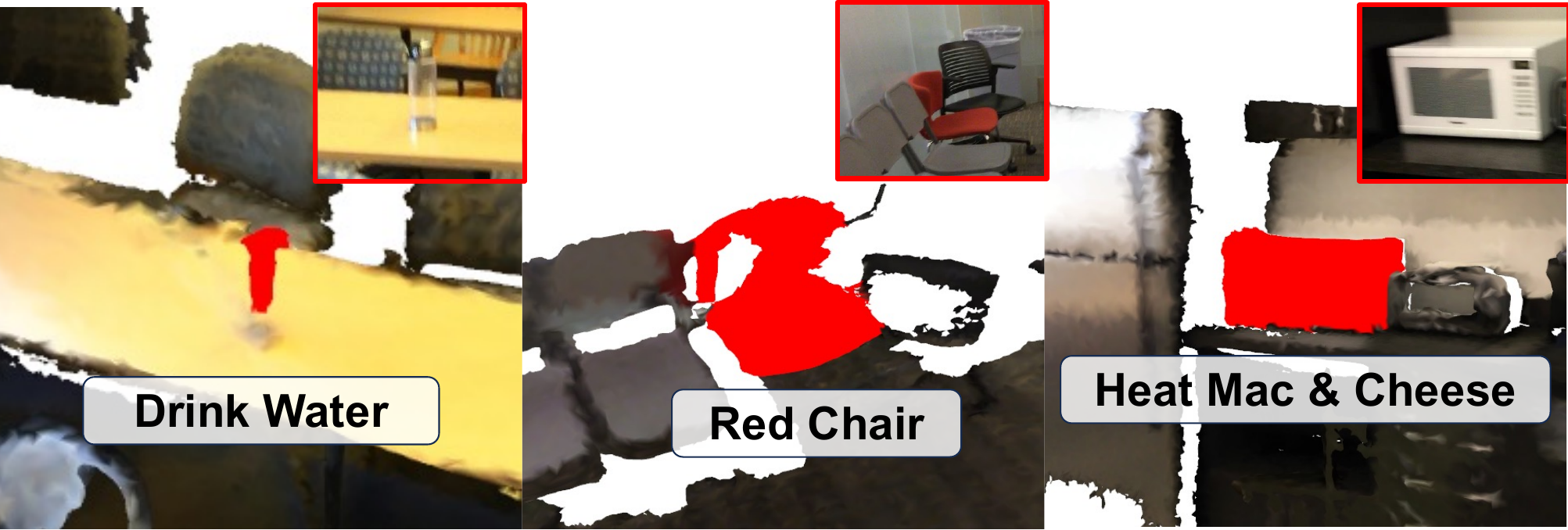}
  \vspace{-0.3cm}
  \caption{\textbf{Examples of open-vocabulary predictions from our method in the ScanNet200 dataset~\cite{scannet}.}
  Our method effectively retrieves instances based on functional descriptions (\textit{e.g.}, drink water, heat mac \& cheese) and object attributes (\textit{e.g.}, red chair).}
  \vspace{-0.7cm}
  \label{fig:exp_openvocab}
\end{figure}

\vspace{-0.2cm}
The task of OV-3DIS~\cite{openmask3d, open3dis, openyolo3d, openins3d, pla, lowis3d, maskclustering, sai3d, regionplc} aims to predict 3D masks for individual objects in a 3D point cloud scene given open-vocabulary text queries (Fig.~\ref{fig:exp_openvocab}). 
OV-3DIS has diverse applications across domains, such as robotics, augmented reality, scene understanding, and 3D visual search. For example, in robotic tasks like indoor navigation and object manipulation, interpreting open-vocabulary queries and localizing corresponding objects in a 3D environment are crucial for effective performance.

Efforts have been made to tackle the task of OV-3DIS.
A two-staged paradigm has been widely adopted across various works~\cite{open3dis, openmask3d, maskclustering, openyolo3d}.
They first generate the class-agnostic 3D proposals and then classify the predicted proposals into open-vocabulary queries.
While some works~\cite{openmask3d, openyolo3d} directly generate 3D proposals from point clouds using pretrained 3D networks~\cite{mask3d, isbnet}, other approaches~\cite{open3dis, maskclustering, sai3d} generate 3D proposals from images.
For 3D proposal generation from images, they leverage vision foundation models (VFMs)~\cite{ovseg, detic, grounded_sam} to ground object regions in each image frame.
Each object region is then lifted to 3D point clouds and temporally aggregated across frames to find complete 3D masks.

While there have been various efforts based on this scheme, e.g., agglomerative clustering~\cite{open3dis}, progressive region growing~\cite{sai3d}, and graph clustering~\cite{maskclustering}, these individual concepts are not mutually exclusive but complementary.
This paper carefully combines the concepts and refines each step to address key challenges, achieving state-of-the-art (SoTA) performance in existing benchmarks.
We generate 3D proposals from both images (i.e., image-based 3D proposals) and aggregated point clouds (i.e., point cloud-based 3D proposals). 
Image-based 3D proposal generation~\cite{open3dis, ovir, sai3d} involves many design choices in three steps: 1) frame-wise 2D object grounding, 2) lifting 2D predictions to 3D point clouds, and 3) 3D proposal aggregation across frames to find complete 3D masks.
Finally, CLIP-based models~\cite{clip} are used to classify the 3D proposals~\cite{open3dis, openmask3d, maskclustering}.

While we adopt this general paradigm, we refine each stage to effectively handle \emph{missing details} in the existing literature.
Also, we devise an additional iterative merging and removal step at the end of the proposal generation to suppress overlapped or partial proposals.

\noindent\textbf{2D Object Grounding.} We observe two representative types of wrong object predictions from VFMs: masks covering multiple objects and partial masks.
While partial masks can be mitigated in later steps by merging or removal, wrong masks covering multiple instances can hardly be separated into individual instances.
Thus, we sort 2D predictions in each frame by their size and remove the overlapped regions from the larger ones to minimize such cases.

\noindent\textbf{2D to 3D Lifting.}
Following existing works~\cite{open3dis, sai3d, sam3d}, we use 3D superpoints~\cite{superpoints} as a basic unit of point cloud operations.
We aim to find a set of 3D superpoints corresponding to each 2D instance in the lifting step.
We adopt two concepts from existing work~\cite{open3dis}: frame-wise and instance-wise visibility scores to remove unconfident superpoints.

\noindent\textbf{Tracking-based 3D Proposal Aggregation.}
We progressively enlarge the lifted 3D superpoints of instances by tracking them sequentially, analogous to OVIR-3D~\cite{ovir}.
However, ours has unique features to improve the limitations of existing works. 
First, we adopt a superpoint-level intersection over union (sIOU) metric instead of a point-level IOU.
This effectively reduces memory usage and computation time.
Also, we apply frame-wise sIOU comparison to match a new observation to existing tracklets (i.e., a list of tracked 2D instances and their lifted 3D superpoints).
Specifically, we compare a new observation with each tracked instance in tracklets to find a match.
We observe that such frame-wise comparisons induce robustness to wrong 2D predictions and noisy projections compared to tracklet-wise comparisons~\cite{ovir} (i.e., using a representative 3D mask for each tracklet by aggregating 3D superpoints of tracked instances).

\noindent\textbf{Iterative Merging/Removal.} We suppress overlapped or partial proposals by merging and removing them.
We iteratively merge proposals if they have large overlaps.
We refine merged proposals using multi-view consensus~\cite{maskclustering, ovir} after every merge iteration.
After the merging step, we remove partial masks if they are included in other proposals.

\noindent\textbf{Instance Classification.} 
We classify the aggregated 3D proposals into open-vocabulary queries.
While existing works~\cite{open3dis, openmask3d} leverage CLIP~\cite{clip} for classification, they can be contaminated by co-visible objects or be sensitive to irregularly shaped objects.
Instead, we adopt Alpha-CLIP~\cite{alphaclip} to obtain an object-centric representation by attending object regions using alpha-channel masks.
Additionally, we introduce a Standardized Maximum Similarity (SMS) score as a proxy for uncertainty to reduce false positives.
The maximum similarity score is standardized using scene-specific statistics, and proposals with low SMS scores are removed from classification.
Such a classification strategy helps enhance precision, delivering SoTA performance on benchmarking datasets.

\begin{figure*}
  \centering
  \includegraphics[width=\linewidth]{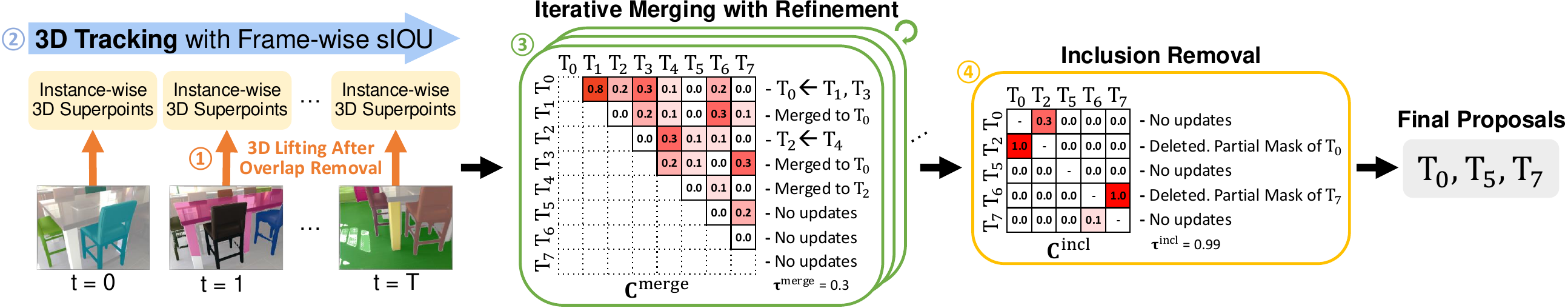}
  \vspace{-0.41cm}
  \caption{\textbf{Overview of image-based 3D proposal generation.}
  We first remove overlaps between 2D predictions within each frame and lift them to 3D point cloud using a camera projection matrix.
  Afterward, we aggregate 3D-lifted predictions across frames using a frame-wise sIOU metric with tracking.
  These 3D proposals are further iteratively merged and refined to progressively merge similar proposals based on a predefined threshold ($\tau^\text{merge}$).
  At last, we remove partial masks if their inclusion ratios to other proposals are higher than a predefined threshold ($\tau^\text{incl}$).
  For further details about merging and removal, see the last paragraph of Sec.~\ref{subsec:proposal}.}
  \vspace{-0.6cm}
  \label{fig:method_overview}
\end{figure*}

Our contributions are summarized as follows:
\begin{itemize}
    \item We carefully combine the existing concepts and refine 3D proposal generation by removing overlaps in 2D predictions and applying robust 3D tracking for aggregation.
    \item We introduce an additional iterative merging/removal step after aggregation to suppress false positives coming from overlapped or partial 3D proposals.
    \item We take advantage of object-centric feature representation by replacing CLIP with Alpha-CLIP and further reduce false positive 3D proposals by measuring the Standardized Maximum Similarity (SMS) score.
    \item We demonstrate significant improvements over SoTA methods on ScanNet200 and S3DIS datasets across AP and AR metrics.
\end{itemize}

%% file: sec/2_related_work.tex
\section{Related Work}
\label{sec:related_work}
\subsection{Closed-vocabulary 3D Instance Segmentation}
This task aims to predict 3D instance segmentation masks by assuming a closed set of classes.
Several methods~\cite{bbox_f3dis1, bbox_f3dis2, bbox_f3dis3, bbox_f3dis4, bbox_f3dis5} have proposed to predict bounding boxes and segment out the instance in each of the bounding boxes.
Another group of approaches~\cite{cluster_f3dis, cluster_f3dis2, cluster_f3dis3, cluster_f3dis4, cluster_f3dis5, cluster_f3dis6, cluster_f3dis7, cluster_f3dis8} builds the instances from point embeddings by using graphs or clustering algorithms.
Lastly, the most recent line of work~\cite{mask3d, isbnet, dyco3d, oneformer3d} adopts transformer architecture~\cite{transformer} or dynamic convolution~\cite{dynamic_conv1, dynamic_conv2} to predict the 3D instance proposals from the point cloud.
Mask3D~\cite{mask3d} utilizes transformer architecture along with sparse convolution, demonstrating state-of-the-art performance.
Another work, ISBNet~\cite{isbnet}, proposes to use improved kernel generation and bounding-box-guided dynamic convolutions.
In our paper, Mask3D and ISBNet are used as our 3D instance segmentation networks, while other networks are also applicable to ours.

\subsection{Open-vocabulary 2D Grounding}
One limitation of closed-vocabulary studies is that they hardly generalize to new environments since they cannot identify novel classes that did not appear in the training set.
Addressing such concerns, open-vocabulary 2D grounding aims to identify novel classes by adopting 2D foundation models or adapting a model to new scenes to discover novel classes.
Three different categories exist under this task: open-vocabulary object detection~\cite{yoloworld, detic, grounding_dino, ovod1, ovod2, ovod3, ovod4, ovod5, ovod6}, open-vocabulary semantic segmentation~\cite{ovss1, ovss2, lseg, openseg, ovseg}, and open-vocabulary instance segmentation~\cite{grounded_sam, seem, odise}.
Most works~\cite{ovseg, vild, openseg, lseg, detic, yoloworld} propose to align their representations to those of pre-trained vision-language models (VLMs) such as CLIP~\cite{clip}.
In our work, we utilize Grounded SAM~\cite{grounded_sam} as our 2D instance grounding method, which utilizes both Grounding DINO~\cite{grounding_dino} and Segment-Anything Model (SAM)~\cite{sam}.
Grounding DINO detects object bounding boxes with given open-vocabulary queries, and SAM predicts the instance mask in each bounding box.

\subsection{Open-vocabulary 3D Instance Segmentation}
This task aims to address the open-vocabulary grounding problems in 3D point clouds.
Several studies~\cite{openscene, lerf, conceptfusion, SA3Dobj, cg3d} have proposed to align the point embeddings to the CLIP embeddings.
However, they often require clustering algorithms to find the instances, heavily relying on the accuracy of clustering algorithms.
Also, some of them are trained on specific datasets, which may limit their generalization. 
On the other hand, other studies~\cite{openmask3d, open3dis, maskclustering, openyolo3d, openins3d} adopt a two-stage scheme, which first generates class-agnostic masks and then classifies the instances.
OpenMask3D~\cite{openmask3d} utilizes Mask3D~\cite{mask3d} to generate the class-agnostic instances in the 3D point cloud and project the instances to 2D images to extract their CLIP embeddings.
Also, OpenYOLO3D~\cite{openyolo3d} classifies 3D proposals generated by Mask3D~\cite{mask3d} using open-vocabulary 2D object detector~\cite{yoloworld}.
However, while they demonstrate promising results, using a pre-trained 3D instance segmentation model often fails at detecting novel or ``tail'' classes since they do not or rarely appear during training.
Thus, OVIR-3D~\cite{ovir} and SAI3D~\cite{sai3d} utilize image-based 3D proposals as an alternative, and Open3DIS~\cite{open3dis} uses both image-based and point cloud-based proposals to improve the recall of tail classes.

%% file: sec/3_method.tex
\vspace{-0.3cm}
\section{Method}
\label{sec:method}
\vspace{-0.3cm}
The task of OV-3DIS is to predict a list of 3D instance masks $\mathbf{m} \in \{0, 1\}^{K\times N}$ that correspond to a list of user queries $\mathcal{Q}$ from a sequence of images $\mathcal{I}$ and point cloud $\mathbf{P}\in\mathbb{R}^{N\times3}$.
$K$ denotes the number of 3D proposals, and $N$ denotes the number of points in the point cloud. 
We generate proposals from both images and point clouds.
Our image-based proposal generation is composed of four steps: 2D object grounding, 2D-to-3D lifting, 3D proposal aggregation, and iterative merging/removal.
Fig.~\ref{fig:method_overview} illustrates the latter three steps in detail with examples.
For point cloud-based 3D proposals, we utilize pre-trained 3D instance segmentation models~\cite{mask3d, isbnet} and discard the class predictions, retaining only the class-agnostic masks. 
We concatenate the proposals from both modalities and classify them into one of the open-vocabulary queries using Alpha-CLIP with SMS-based filtering.
We will present our image-based proposal generation in Sec.~\ref{subsec:proposal}, and then elaborate on instance classification in Sec.~\ref{subsec:classification}.

\subsection{Image-based Proposal Generation}
\label{subsec:proposal}
Leveraging VFMs~\cite{grounded_sam, sam, grounding_dino}, image-based proposals provide a complementary approach for detecting novel classes not covered during the training of the 3D instance segmentation models~\cite{mask3d, isbnet}.
To generate these proposals, we: 
1) ground 2D objects and remove overlapping regions, 
2) lift 2D predictions to 3D superpoints,
3) aggregate 3D proposals over frames using tracking,
4) refine 3D proposals,
5) iteratively merge and remove redundant proposals.

\noindent\textbf{2D Object Grounding and Overlap Removal.}
We use Grounded SAM~\cite{grounded_sam} to segment 2D instance masks in each image from open-vocabulary queries.
For each frame, predicted 2D instance masks are sorted by their size, and overlapping regions of larger masks are removed (i.e., overlap removal).
This step mitigates the issue of masks frequently capturing multiple objects, which can lead to 3D proposals spanning multiple instances. Although overlap removal may result in partial 3D proposals, we found that separating 3D masks containing multiple instances into distinct masks is far more challenging than starting with multiple partial 3D masks for each instance and merging/removing them after aggregation.

\begin{figure}
  \centering
  \includegraphics[width=0.70\linewidth]{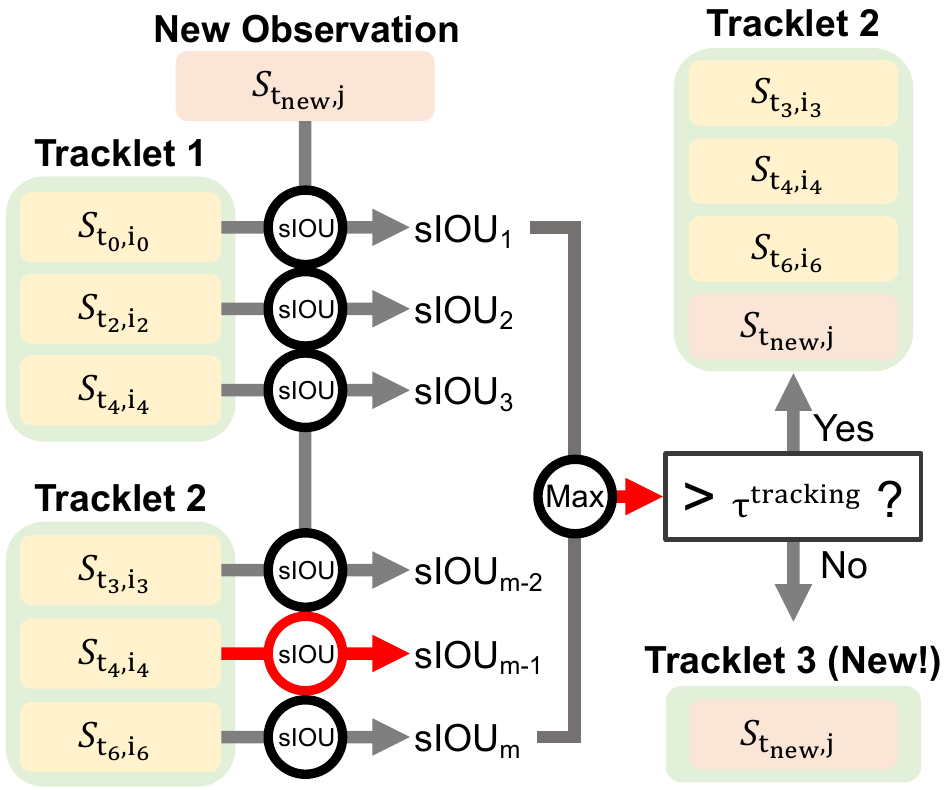}
  \vspace{-0.2cm}
  \caption{\textbf{Matching tracklets with a new observation.}
  We conduct frame-wise sIOU comparisons between a new observation and each tracked instance in tracklets.
  If the maximum sIOU exceeds a predefined threshold, we update the corresponding tracklet; otherwise, a new tracklet is initialized.
  }
  \vspace{-0.6cm}
  \label{fig:method_matching}
\end{figure}

\noindent\textbf{2D Instance to 3D Superpoints Lifting.}
For 3D point cloud lifting of 2D pixels, we leverage camera matrices, i.e., the multiplication of an intrinsic and extrinsic matrix for each frame.
We adopt two concepts from Open3DIS~\cite{open3dis}: frame-wise visibility ratio $r_t(\mathbf{s})$ and instance-wise visibility ratio $c_{t, \cdot}(\mathbf{s})$ for filtering out superpoints.

Specifically, $r_t(\mathbf{s})$ denotes the visibility ratio of superpoint $\mathbf{s}$ with respect to image $\mathbf{I}_t$.
This ratio is defined as the proportion of 3D points within the superpoint whose projections are visible in the image. 
Similarly, $c_{t,i}(\mathbf{s})$ indicate the ratio of visible superpoint $\mathbf{s}$ supported (\textit{i.e.,} overlapped) by 2D mask of the $i$-th instance in image $\mathbf{I}_t$.
This is defined as the proportion of visible 3D points within the instance mask relative to the total number of visible 3D points in the image.
Based on these definitions, we define two sets of 3D superpoints: one for 3D superpoints visible in the image and another for 3D superpoints visible within a specific instance mask:
\begin{equation}
    \begin{split}
        \mathcal{S}_t &= \{\mathbf{s} \mid r_t(\mathbf{s}) > \tau^\text{img} \}\\
        \mathcal{S}_{t, i} &= \{\mathbf{s} \mid r_t(\mathbf{s}) > \tau^\text{img} \text{ and } c_{t, i}(\mathbf{s}) > \tau^{\text{inst}}\},
    \end{split}
\end{equation}
where $\tau^\text{img}$ and $\tau^{\text{inst}}$ are predefined thresholds for filtering out 3D superpoints with low visibility and/or support from the $i$-th 2D segment.
This formulation ensures that only 3D superpoints with sufficient visibility and/or support are included for further processing. 

\noindent\textbf{Tracking-based 3D Proposal Aggregation.}
We aggregate 2D instance masks and their corresponding 3D superpoints by tracking them over frames. 
We maintain a list of tracklets, where each tracklet records a list of tracked 2D instance masks and their lifted 3D superpoints. Note that each tracklet corresponds to a single 3D instance proposal after aggregation.

Tracklets are initialized using the lifted 3D superpoints of 2D instances from the first image.
Afterward, we associate new observations from the next frames with existing tracklets using frame-wise sIOU metrics.
Specifically, we compute the sIOU between the lifted 3D superpoints of the new observation and each tracked instance in each tracklet (see Fig.~\ref{fig:method_matching}).
If the highest sIOU exceeds a predefined threshold $\tau^\text{tracking}$, the new observation is assigned to the corresponding tracklet for update.
Otherwise, a new tracklet is created for this new instance.
When measuring sIOU between two sets, we only consider co-visible superpoints in both image frames.
Formally noting, given the $i$-th instance mask from an image $\mathbf{I}_{t_a}$ and the $j$-th instance mask from $\mathbf{I}_{t_b}$, we denote their corresponding 3D superpoints within each instance mask as $\mathcal{S}_{t_a, i}$ and $\mathcal{S}_{t_b, j}$ respectively and a co-visible 3D superpoints set between images as $\text{Vis}_{t_a,t_b}= \mathcal{S}_{t_a} \cap \mathcal{S}_{t_b}$.
sIOU between these two instance masks from two frames is defined as:
\begin{equation}
\text{sIOU} = \frac{|(\mathcal{S}_{t_a, i} \cap \mathcal{S}_{t_b, j})|}{|(\mathcal{S}_{t_a, i} \cup \mathcal{S}_{t_b, j}) \cap \text{Vis}_{t_a,t_b}|},
\end{equation}
where $|\cdot|$ denotes the cardinality.

\begin{figure}[t]
  \centering
  \includegraphics[width=0.85\linewidth]{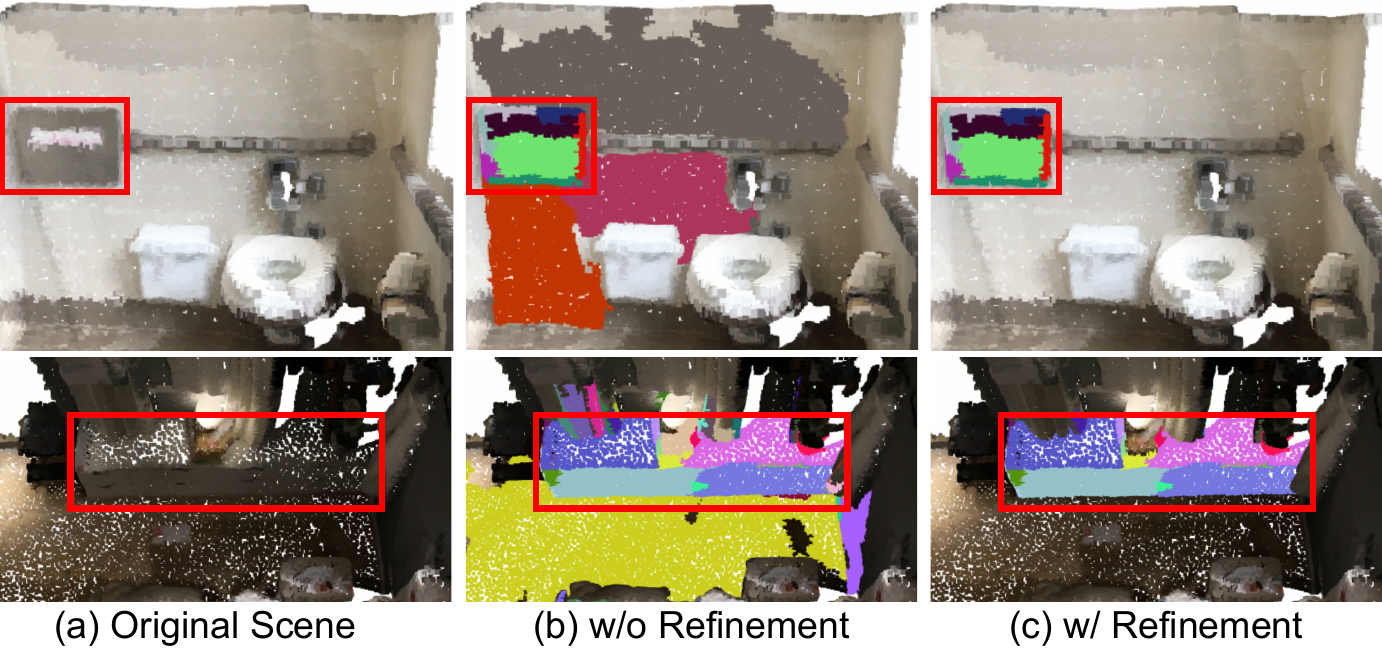}
  \vspace{-0.3cm}
  \caption{\textbf{Effectiveness of 3D proposal refinement.} Red boxes indicate the object of interest, and segments of different colors denote 3D superpoints. Without refinement, the 3D instance proposal often extends beyond the object boundaries due to noisy 2D-to-3D projections or inaccurate mask predictions. With refinement, irrelevant 3D superpoints are removed, and our method successfully removes 3D superpoints that do not belong to the object, resulting in geometrically consistent and precise predictions.
  }
  \vspace{-0.6cm}
  \label{fig:exp_refinement}
\end{figure}

\noindent\textbf{3D Proposal Refinement.}
After tracking, we have an additional refinement step for the 3D proposal in each tracklet by removing 3D superpoints that are infrequently visible in the tracked 2D instances across multiple views.
We adopt a concept from MaskClustering~\cite{maskclustering} and OVIR-3D~\cite{ovir} and calculate a \emph{superpoint-level} multi-view consensus rate.
As illustrated in Fig.~\ref{fig:exp_refinement}, removing superpoints with low visibility effectively refines the proposal to have a tight, semantic-aligned boundary.
For each superpoint in a tracklet, the multi-view consensus rate is defined as the ratio of tracked frames in which the superpoint appears within the instance mask to the total number of frames where it is visible. 3D superpoints with a consensus rate below a predefined threshold \(\tau^\text{ref}\)
are removed from the tracklet, ensuring only reliable superpoints are retained.

\begin{figure}[t]
  \centering
  \includegraphics[width=0.9\linewidth]{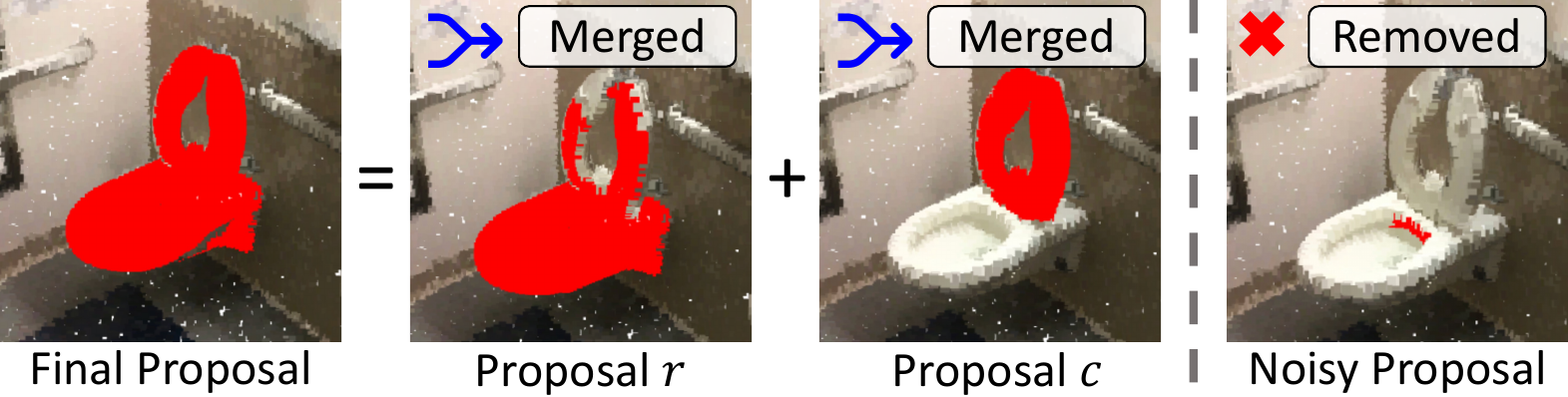}
  \vspace{-0.4cm}
  \caption{\textbf{Visualization of merged and removed proposals in the ScanNet200 dataset.} Overlapping and noisy proposals often emerge after instance tracking. We effectively handle these issues by merging duplicate proposals and eliminating noisy ones, ensuring high-quality proposals.}
  \vspace{-0.6cm}
  \label{fig:method_merge}
\end{figure}

\noindent\textbf{Iterative 3D Proposal Merge and Removal.}
While our overlap removal step in the 2D grounding step effectively removes masks spanning multiple instances, it may generate partial masks of instances.
For example, as shown in Fig.~\ref{fig:method_merge}, a single object may be decomposed into multiple partial 3D proposals, each covering only part of the object.
To address this, we merge these partial proposals into a complete 3D representation.
Moreover, we apply this merging iteratively so that we can progressively enlarge instances at each iteration.
Also, each merge is followed by proposal refinement using multi-view consensus to exclude noisy superpoints from later merging.
Conversely, when a larger 3D proposal contains smaller, redundant proposals, we remove the redundant ones to ensure higher precision.

Suppose we have $K$ tracklets, each with a 3D mask proposal represented as $\mathbf{m}_k \in \{0, 1\}^{N}$, derived from the tracked 3D superpoints within each tracklet.
For each merging iteration, we compute IOU between a pair of 3D proposals, constructing a cost matrix $\mathbf{C}^\text{merge} \in [0, 1]^{K \times K}$ that is a strictly upper-triangular matrix.
For each 3D proposal, we identify other proposals with an IOU exceeding a predefined threshold $\tau^\text{merge}$ and merge them into the current 3D proposal.
This process is repeated until no further merges are possible.
After each 3D proposal merge, we also merge their corresponding tracklets and refine the resulting 3D proposal using the multi-view consensus rate. More detailed implementation can refer to supplementary materials.

After merging proposals, we remove smaller 3D proposals that are contained within larger ones.
Given two 3D proposals $\mathbf{m}_r, \mathbf{m}_c \in \{0, 1\}^N$, we define an inclusion rate of $\mathbf{m}_r$ within $\mathbf{m}_c$ as $r^\text{incl}(\mathbf{m}_{r}, \mathbf{m}_{c})$, which is the proportion of $\mathbf{m}_r$ included in $\mathbf{m}_c$, with a value of 1 indicating that $\mathbf{m}_r$ is fully contained within $\mathbf{m}_c$.
Using this ratio, we construct an inclusion cost matrix $\mathbf{C}^\text{incl} \in [0, 1]^{K \times K}$, which is a full matrix since the inclusion ratio is asymmetric.
For each 3D proposal, if its inclusion rate with respect to any other proposal exceeds a predefined threshold $\tau^\text{incl}$, the proposal is removed. Unlike the merge process, this filtering step is applied only once.

\subsection{Open-Vocabulary Instance Classification}
\label{subsec:classification}
For instance classification, we concatenate class-agnostic proposals from image-based and point cloud-based methods and apply non-maximum suppression (NMS) with an IOU threshold of 0.95.
We prioritize the point cloud-based proposals over the image-based ones since they have fewer false positives.

\begin{figure}[t]
  \centering
  \includegraphics[width=0.6\linewidth]{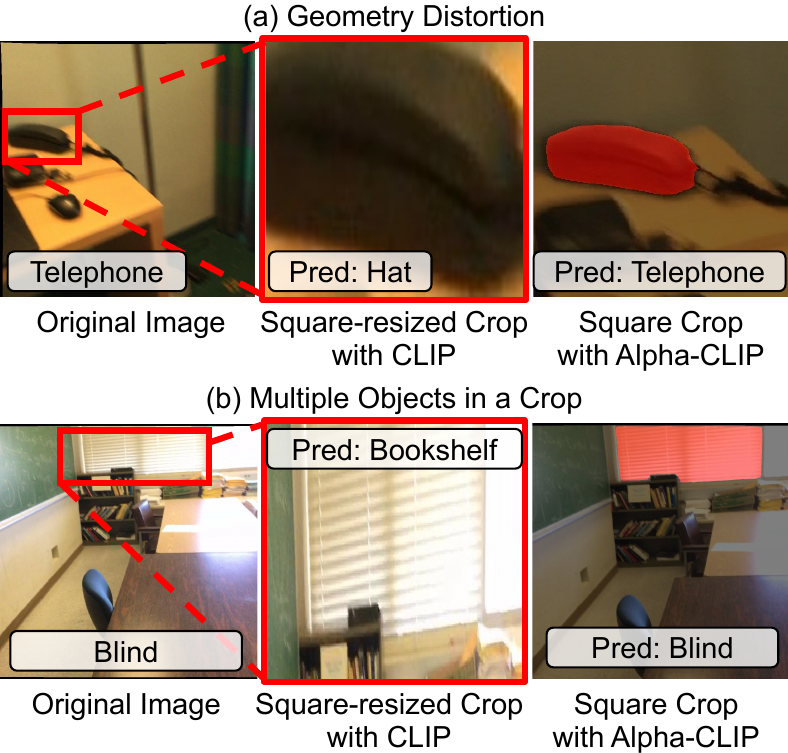}
  \vspace{-0.3cm}
  \caption{\textbf{Failure cases of using CLIP for instance classification.}
  CLIP fails when the shape of the object gets distorted or when other objects are also present within the crop.}
  \vspace{-0.7cm}
  \label{fig:method_alphaclip}
\end{figure}

\noindent\textbf{Feature Extraction.}
Previous work~\cite{openmask3d, open3dis} leverages CLIP~\cite{clip} to extract visual features from cropped image regions using projected instance bounding boxes. However, this approach has notable limitations (Fig.~\ref{fig:method_alphaclip}):
(a) Resizing the crop to a square aspect ratio distorts the object’s original geometry, hindering CLIP’s ability to capture its geometric characteristics.
(b) Visual features are contaminated by co-visible objects (e.g., bookshelves, tables), leading to poor predictions from CLIP.
To address these issues, we adopt Alpha-CLIP~\cite{alphaclip} to enforce an object-centric focus to the model.
Alpha-CLIP incorporates object masks as an additional input to guide the model's attention.
The object masks are generated using SAM by projecting the predicted 3D proposals onto images and querying with bounding boxes or subsampled points.
We also apply a square crop during preprocessing to preserve the object’s geometry while ensuring compatibility with the model’s input requirements.

\begin{table*}[ht!]
\begin{center}
\footnotesize
\scalebox{0.8}{
\begin{tabular}{cccccccccc}
\toprule
\multirow{2}{*}{Eval. Protocol} & \multirow{2}{*}{Methods} & \multicolumn{2}{c}{3D Proposals} & \multirow{2}{*}{mAP} & \multirow{2}{*}{mAP$_{50}$} & \multirow{2}{*}{mAP$_{25}$} & \multirow{2}{*}{mAP$_\text{head}$} & \multirow{2}{*}{mAP$_\text{common}$} & \multirow{2}{*}{mAP$_\text{tail}$} \\
&& Image-based & Point cloud-based & & & & & & \\
\drule
\multirow{2}{*}{Fully Supervised}& ISBNet~\cite{isbnet} & \textcolor{red}{\xmark} & \textcolor{forestgreen}{\checkmark} & 24.5 & 32.7 & 37.6 & 38.6 & 20.5 & 12.5 \\
& Mask3D~\cite{mask3d} & \textcolor{red}{\xmark} & \textcolor{forestgreen}{\checkmark} & 26.9 & 36.2 & 41.4 & 39.8 & 21.7 & 17.9 \\
\drule
\multirow{10}{*}{Top-1}& SAM3D$^\dagger$~\cite{sam3d}              & \textcolor{forestgreen}{\checkmark} & \textcolor{red}{\xmark} & 9.8 & 15.2 & 20.7 & 9.2 & 8.3 & 12.3 \\
& OVIR-3D$^\dagger$~\cite{ovir}            & \textcolor{forestgreen}{\checkmark} & \textcolor{red}{\xmark} & 9.3 & 18.7 & 25.0 & 9.8 & 9.4 & 8.5 \\
& SAI3D$^\dagger$~\cite{sai3d}  & \textcolor{forestgreen}{\checkmark} & \textcolor{red}{\xmark} & 12.7 & 18.8 & 24.1 & 12.1 & 10.4 & 16.2 \\
& \textbf{Ours (2D Only)} & \textcolor{forestgreen}{\checkmark} & \textcolor{red}{\xmark} & \textbf{21.5} & \textbf{31.2} & \textbf{37.7} & \textbf{18.8} & \textbf{19.6} & \textbf{26.9} \\
\cmidrule{2-10}
& OpenIns3D~\cite{openins3d} & \textcolor{red}{\xmark} & \textcolor{forestgreen}{\checkmark} & 8.8 & 10.3 & 14.4 & 16.0 & 6.5 & 4.2 \\
& OpenMask3D~\cite{openmask3d} & \textcolor{red}{\xmark} & \textcolor{forestgreen}{\checkmark} & 15.4 & 19.9 & 23.1 & 17.1 & 14.1 & 14.9 \\
&OpenYOLO3D~\cite{openyolo3d} & \textcolor{red}{\xmark} &  \textcolor{forestgreen}{\checkmark} & 21.9 & 28.3 & 31.7 & 25.6 & 21.1 & 18.5 \\
& \textbf{Ours (3D Only)} & \textcolor{red}{\xmark} & \textcolor{forestgreen}{\checkmark} & \textbf{24.2} & \textbf{31.8} & \textbf{36.4} & \textbf{27.2} & \textbf{22.3} & \textbf{23.1} \\
\cmidrule{2-10}
& OpenScene~\cite{openscene} & \textcolor{forestgreen}{\checkmark}  &  \textcolor{forestgreen}{\checkmark} & 11.7 & 15.2 & 17.8 & 13.4 & 11.6 & 9.9 \\
& \textbf{Ours (2D + 3D)} & \textcolor{forestgreen}{\checkmark} & \textcolor{forestgreen}{\checkmark} & \textbf{25.8} & \textbf{32.5} & \textbf{36.2} & \textbf{26.3} & \textbf{23.2} & \textbf{28.2} \\
\drule
\multirow{8}{*}{Top-K} & Open3DIS~\cite{open3dis}  & \textcolor{forestgreen}{\checkmark} & \textcolor{red}{\xmark} & 18.2 & 26.1 & 31.4 & 18.9 & 16.5 & 19.2 \\
& \textbf{Ours (2D Only)} & \textcolor{forestgreen}{\checkmark} & \textcolor{red}{\xmark} & \textbf{25.4} & \textbf{37.4} & \textbf{44.4} & \textbf{23.4} & \textbf{23.5} & \textbf{30.2} \\
\cmidrule{2-10}
& Open3DIS~\cite{open3dis} & \textcolor{red}{\xmark} & \textcolor{forestgreen}{\checkmark} & 18.6 & 23.1 & 27.3 & 24.7 & 16.9 & 13.3 \\
& OpenYOLO3D~\cite{openyolo3d} & \textcolor{red}{\xmark} &  \textcolor{forestgreen}{\checkmark} & 24.7 & 31.7 & 36.2 & 27.8 & 24.3 & 21.6 \\
& \textbf{Ours (3D Only)} & \textcolor{red}{\xmark} & \textcolor{forestgreen}{\checkmark} & \textbf{29.0} & \textbf{37.6} & \textbf{42.8} & \textbf{33.0} & \textbf{28.1} & \textbf{25.3} \\
\cmidrule{2-10}
& Open3DIS~\cite{open3dis} & \textcolor{forestgreen}{\checkmark}  &  \textcolor{forestgreen}{\checkmark} & 23.7 & 29.4 & 32.8 & 27.8 & 21.2 & 21.8 \\
& \textbf{Ours (2D + 3D)} & \textcolor{forestgreen}{\checkmark} & \textcolor{forestgreen}{\checkmark} & \textbf{32.7} & \textbf{41.4} & \textbf{45.3} & \textbf{34.5} & \textbf{30.7} & \textbf{33.1} \\
\bottomrule
\end{tabular}}
\end{center}
\vspace*{-0.6cm}
\caption{\textbf{OV-3DIS results on the ScanNet200 validation set~\cite{scannet}.}
Top-1 evaluation protocol refers to assigning one predicted class per instance mask, and Top-K evaluation protocol~\cite{open3dis, openyolo3d} refers to allowing multiple predicted classes per instance mask.
We evaluate methods under three settings: image-based 3D proposals only (\textit{i.e.,} 2D only), point cloud-based 3D proposals only (\textit{i.e.,} 3D only), and a combination of both (\textit{i.e.,} 2D+3D).
In all three settings across different protocols, our method achieves the SoTA performance, significantly outperforming other methods.
$^\dagger$numbers are adopted from SAI3D~\cite{sai3d}.
}
\label{tab_scannet}
\vspace*{-0.6cm}
\end{table*}

We adopt a similar approach to OpenMask3D~\cite{openmask3d} for visual embedding extraction. Given a 3D proposal and the visual encoder from Alpha-CLIP, we project the proposal onto all 2D images and select a subset of images with the highest visibility for multiscale visual feature extraction. Let $\mathbf{f}_{v,k}^l$ represent the CLIP feature extracted at a scale level $l$ from the $v$-th image for the $k$-th 3D proposal. The final L2-normalized feature $\mathbf{F}_k$ for this proposal is computed as:
\begin{equation}
    \mathbf{F}_k = \sum\limits_{v \in \mathcal{V}_k} \sum\limits_{l \in \mathcal{L}} \mathbf{f}_{v,k}^l \cdot \alpha_{v,k},
\end{equation}
where $\mathcal{V}_k$ denotes the set of images with top visibility for the $k$-th 3D proposal, and $\alpha_{v,k} \in [0, 1]$ is the visibility ratio of the $k$-th 3D proposal in image $\mathbf{I}_v$. This ratio is defined as the number of visible points in the image divided by the total number of points in the 3D proposal.
Finally, given $K$ proposals and $C$ text queries, we compute the cosine similarity between the visual features of the proposals and the text features, resulting in a similarity matrix $\mathbf{L} \in [-1, 1]^{K \times C}$.

\noindent\textbf{3D Proposal Filtering with SMS.}
We further suppress unconfident proposals by using the CLIP similarity score as a proxy for uncertainty.
However, CLIP scores are not normalized across different text embeddings, making it challenging to apply a single filtering threshold for all queries.
To address this, we standardize the maximum similarity scores (i.e., SMS score) within each text embedding to obtain relative scores.
Specifically, for each query $q_c$, we compute the mean and variance of the similarity scores as $\mu_c = \frac{1}{K} \sum_k \mathbf{L}_{k,c}$ and $\sigma_c^2 = \frac{1}{K} \sum_k (\mathbf{L}_{k,c} - \mu_c)^2$, where $K$ is the total number of proposals. Next, for each proposal $k$, we identify the maximum similarity value $\mathbf{L}_{k, c_\text{max}}$ across all queries and standardize it using the corresponding statistics:
$c_k^\text{SMS} = \frac{\mathbf{L}_{k, c_\text{max}} - \mu_{c_\text{max}}}{\sigma_{c_\text{max}}}.$
Proposals with an SMS score below a predefined threshold $\tau^\text{SMS}$ are removed from the predictions, ensuring more reliable filtering.

%% file: sec/4_experiments.tex
\vspace{-0.2cm}
\section{Experiments}
\vspace{-0.2cm}
\label{sec:exp}
\subsection{Experimental Setup}
\noindent\textbf{Datasets.}
We evaluate our method on three datasets: ScanNet200~\cite{scannet}, S3DIS~\cite{s3dis}, and Replica~\cite{replica}.
\textbf{ScanNet200} is a real-world dataset comprising diverse indoor environments with 200 object categories. It includes 1,201 scenes in the training set and 312 scenes in the validation set. Object categories are divided into head, common, and tail classes based on their frequency. We validate our method and baselines on the validation set, reporting performance for each category group (\textit{i.e.,} head, common, tail) as well as the overall performance.
\textbf{S3DIS} consists of 271 scenes from 6 different areas, with Area 5 used for our evaluation. Although the dataset includes 13 classes, we exclude ``stuff" categories such as floor, ceiling, wall, and clutter from our evaluation to focus on object-centric performance.
\textbf{Replica} is a synthetic dataset created from digital replicas of real-world scenes, featuring 48 object classes across 8 different scenes. On this dataset, we assess generalization performance by evaluating a 3D instance segmentation model~\cite{mask3d} trained on ScanNet200.

\noindent\textbf{Evaluation Metrics.}
We measure mean average precision (mAP) and mean average recall (mAR) at IOU thresholds of 25\% and 50\%.
Additionally, we measure mAP and mAR across IOU thresholds ranging from 50\% to 95\% with 5\% increments.
For class-agnostic evaluations, we calculate AP and AR on those IOU ranges.

\noindent\textbf{Evaluation Protocols.}
We found existing literature adopts different evaluation strategies. Several works~\cite{openmask3d, sai3d, openins3d, openscene} assign one class prediction per each 3D instance (i.e., Top-1), while other works~\cite{open3dis, openyolo3d} allows multiple predictions per each 3D instance by selecting Top-K predictions (e.g., top 300 / 600) over class predictions of all instances, based on their prediction scores.
For fair comparisons, we evaluate our method on both evaluation settings on ScanNet200 and Replica. For the S3DIS dataset, we only evaluate by using the Top-1 strategy. Further details can be found in the supplementary materials.

\begin{figure*}
  \centering
  \includegraphics[width=0.85\linewidth]{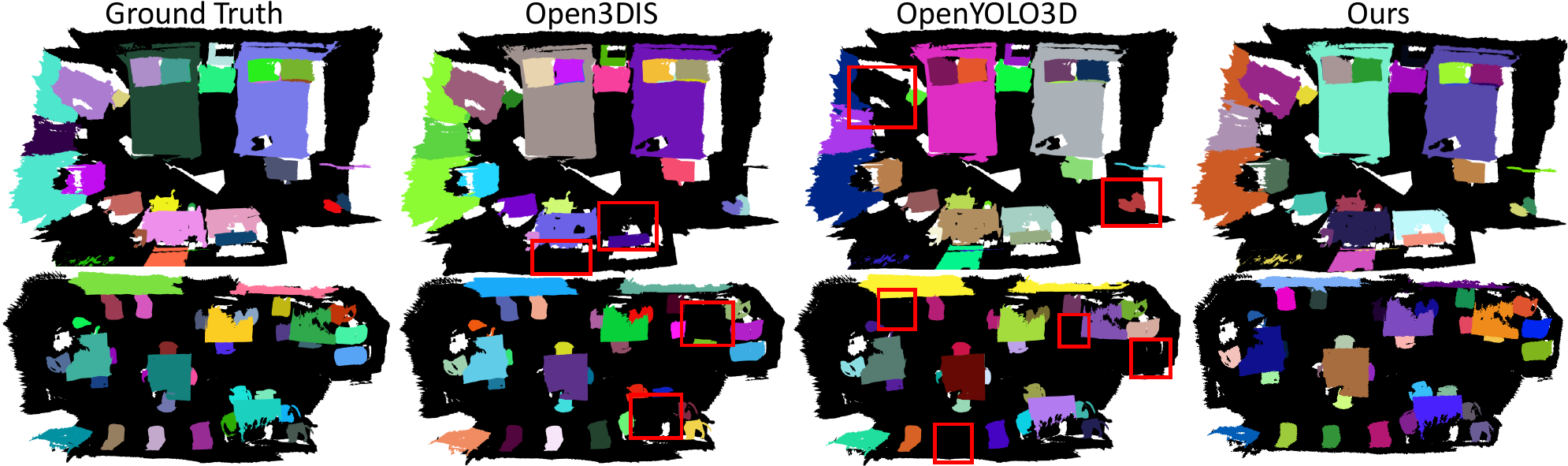}
  \vspace{-0.3cm}
  \caption{\textbf{Qualitative comparisons on the ScanNet200 dataset.} 
  Black regions indicate empty predictions (\textit{no object}), while red boxes highlight objects missed by other methods but successfully detected by ours. 3D instance masks are colored randomly.
  }
  \vspace*{-0.6cm}
  \label{fig:exp_qualitative}
\end{figure*}
\noindent\textbf{Implementation Details.}
For the ScanNet200 dataset, we downsample the number of image frames by a factor of 5 to reduce computational load.
We follow the same setting of OpenMask3D~\cite{openmask3d} for multi-scale CLIP feature extraction, i.e., 3 scale levels with an expansion ratio of 0.2.
We set the following thresholds for both ScanNet200 and S3DIS: $\tau^\text{img}=0.1$, $\tau^\text{inst}=0.3$, $\tau^\text{tracking}=0.3$, $\tau^\text{merge}=0.3$, $\tau^\text{ref}=0.4$, and $\tau^\text{incl}=0.99$. For the Replica dataset, we adjust $\tau^\text{merge}$ to 0.7 and disable multiview consensus ratio-based filtering, as Replica is a synthetic dataset without projection errors.
Additionally, we observe a distribution shift in the CLIP visual representations, as CLIP is trained on real-world data while Replica consists of synthetic data. This shift enlarges the gap between the visual and text embeddings of CLIP. To address this, inspired by prior work on handling distributional gaps~\cite{robustnet, zca, whitening1, WCT, reclip}, we perform dimension reduction by computing the first principal axis of the visual CLIP features and removing its contribution from both visual and text embeddings. We use the template ``\textit{a blurry photo of \{CLASS\_NAME\} in a room.}" Further details are provided in the supplementary materials.

\begin{table}[t]
\begin{center}
\footnotesize
\scalebox{0.86}{
\begin{tabular}{ccccccc}
\toprule
Methods & mAP & mAP$_{50}$ & mAP$_{25}$ & mAR & mAR$_{50}$ & mAR$_{25}$ \\
\drule
Open3DIS$^\dagger$~\cite{open3dis} & 24.5 & 36.2 & 49.3 & 30.5 & 43.4 & 55.3 \\
\textbf{Ours (2D Only)} & \textbf{26.9} & \textbf{41.4} & \textbf{51.9} & \textbf{35.3} & \textbf{52.5} & \textbf{62.5} \\ 
\midrule
Open3DIS$^\dagger$~\cite{open3dis} & 26.6 & 33.5 & 39.2 & 34.2 & 41.7 & 47.4 \\
OpenYOLO3D$^\dagger$~\cite{openyolo3d} & 28.1 & 37.3 & 44.6 & 37.8 & 46.3 & 52.0\\
\textbf{Ours (3D Only)} & \textbf{30.4} & \textbf{39.4} & \textbf{47.4} & \textbf{38.7} & \textbf{47.9} & \textbf{55.1} \\ 
\midrule
Open3DIS$^\dagger$~\cite{open3dis} & 28.9 & 37.0 & 43.1 & 44.1 & 54.5 & 61.4  \\
\textbf{Ours (2D + 3D)} & \textbf{31.3} & \textbf{43.5} & \textbf{50.4} & \textbf{48.2} & \textbf{65.1} & \textbf{72.9} \\ 
\bottomrule
\end{tabular}}
\end{center}
\vspace{-0.7cm}
\caption{\textbf{OV-3DIS results on S3DIS~\cite{s3dis}.}
$^\dagger$numbers are obtained using their official codes.
Top-1 evaluation protocol is used.
}
\label{tab_s3dis}
\vspace{-0.7cm}
\end{table}

\vspace{-0.1cm}
\subsection{Quantitative Results}
\vspace{-0.1cm}
\noindent\textbf{ScanNet200.} We adopt Mask3D~\cite{mask3d} trained on the ScanNet200 training set for point cloud-based proposals. Table~\ref{tab_scannet} shows that our method outperforms all baselines in all three settings, \textit{i.e.,} 2D-only, 3D-only, and 2D+3D across all evaluation protocols.
In 2D-only evaluations, our method outperforms previous SoTA methods, SAI3D and Open3DIS, by 8.8\% and 7.2\% , respectively.
Notably, our image-based method excels at predicting tail classes, where other comparing methods often struggle.
Note that tail classes refer to \emph{less frequent, rare} classes in the training set, not based on their object sizes.
For 3D-only evaluations, we use the same point cloud-based 3D proposals for both our method and OpenYOLO3D.
The resulting 2.3\% and 4.3\% mAP improvements over OpenYOLO3D demonstrate the effectiveness of our classification method. Furthermore, using both image-based and point cloud-based proposals, we outperform the previous SoTA, Open3DIS~\cite{open3dis}, by 9.0\% in mAP.

\noindent\textbf{S3DIS.} We train ISBNet~\cite{isbnet} on Area 1$\sim$4, 6 and adopt its predictions on Area 5 as our point cloud-based proposals for all the baselines.
As reported in Table~\ref{tab_s3dis}, our method consistently outperforms the baselines by a large margin in each experiment setting: 2D-only, 3D-only, and 2D+3D.
Using both image-based and point cloud-based proposals boosts the recall significantly, improving 2D-only and 3D-only methods by 12.9\% and 9.5\%, respectively.
Note that we use thing classes only since our task is 3D instance segmentation.
The results for both stuff and thing classes are available in the supplementary materials.

\noindent\textbf{Replica.}
The results are summarized in Table~\ref{tab_replica}. This experiment aims to assess the generalizability of our method by adopting a ScanNet200-trained 3D instance segmentation model for point cloud-based proposals. Our approach consistently outperforms other methods within the same category under both 2D-only and 2D+3D settings, achieving superior results on both mAP and mAR.
In the 3D-only setting, our method significantly surpasses OpenMask3D and Open3DIS, which adopt CLIP~\cite{clip} for predictions.
However, it lags behind OpenYOLO3D~\cite{openyolo3d} in terms of mAP, which does not use CLIP for instance classification.
We hypothesize that the domain gap between real-world data and synthetic data from Replica may degrade the performance of Alpha-CLIP.
Surprisingly, our 2D-only method achieves higher mAP$_{50}$ and mAP$_{25}$ than 3D only methods where the masks are generated from ScanNet200-trained 3D networks.
This highlights the exceptional generalization capability of our 2D-only approach.
Additionally, our 2D+3D method attains the highest mAR across all settings.

\begin{table}[t]
\begin{center}
\footnotesize
\scalebox{0.76}{
\begin{tabular}{cccccccc}
\toprule
Eval. & Methods & mAP & mAP$_{50}$ & mAP$_{25}$ & mAR & mAR$_{50}$ & mAR$_{25}$\\
\drule
\multirow{7}{*}{Top-1}&OVIR-3D~\cite{ovir} & 11.1 & 20.5 & 27.5 & - & - & - \\
&\textbf{Ours (2D Only)} & \textbf{20.8} & \textbf{32.4} & \textbf{38.5} & 28.5 & 43.1 & 49.9 \\
\cmidrule{2-8}
&OpenMask3D~\cite{openmask3d} & 13.1 & 18.4 & 24.2 & - & - & - \\
&OpenYOLO3D~\cite{openyolo3d} & \textbf{23.7} & \textbf{28.6} & \textbf{34.8} & \textbf{26.6} & \textbf{31.9} & \textbf{38.5} \\
&\textbf{Ours (3D Only)}& 22.0 & 26.7 & 32.5 & \textbf{26.6} & 31.5 & 37.0 \\
\cmidrule{2-8}
&\textbf{Ours (2D + 3D)} & \textbf{22.6} & \textbf{31.7} & \textbf{37.7} & \textbf{33.9} & \textbf{46.5} & \textbf{53.6} \\
\midrule
\multirow{7}{*}{Top-K} &Open3DIS$^\dagger$~\cite{open3dis} & 18.2 & 25.9 & 31.0 & 32.3 & 46.2 & 54.9 \\
&\textbf{Ours (2D Only)} & \textbf{21.6} & \textbf{32.6} & \textbf{39.8} & \textbf{39.6} & \textbf{59.5} & \textbf{71.3} \\
\cmidrule{2-8}
&Open3DIS$^\dagger$~\cite{open3dis} & 16.0 & 19.4 & 23.5 & 29.2 & 35.4 & 42.5 \\
&\textbf{Ours (3D Only)} & \textbf{18.9} & \textbf{24.4} & \textbf{32.1} & \textbf{34.0} & \textbf{43.9} & \textbf{57.5} \\
\cmidrule{2-8}
&Open3DIS$^\dagger$~\cite{open3dis} & 18.4 & 23.8 & 28.2 & 33.0 & 42.6 & 50.0 \\
&\textbf{Ours (2D + 3D)} & \textbf{25.7} & \textbf{34.9} & \textbf{42.3} & \textbf{48.8} & \textbf{66.3} & \textbf{79.7} \\
\bottomrule
\end{tabular}}
\end{center}
\vspace{-0.7cm}
\caption{\textbf{OV-3DIS results on Replica~\cite{replica}.}
$^\dagger$numbers are obtained using their official codes. 
}
\label{tab_replica}
\vspace{-0.7cm}
\end{table}

\vspace{-0.1cm}
\subsection{Qualitative Results}
\vspace{-0.2cm}
Fig.~\ref{fig:exp_qualitative} presents qualitative comparisons on the ScanNet200 dataset. Red boxes indicate instances missed by Open3DIS and OpenYOLO3D, while our method successfully detects all objects. These visual results are consistent with the recall metrics: Open3DIS and OpenYOLO3D achieve the mAR of 43.3\% and 47.7\%, respectively, whereas our method significantly outperforms both with an mAR of 61.4\%.
We also present OV-3DIS results using novel text queries in Fig.~\ref{fig:exp_openvocab}. Our framework effectively retrieves 3D instances based on functional descriptions (\textit{e.g.}, drink water) and object attributes (\textit{e.g.}, red chair).

\begin{table}[ht!]
\begin{center}
\footnotesize
\scalebox{0.8}{
\begin{tabular}{cccccccc}
\toprule
Methods & AP & AP$_{50}$ & AP$_{25}$ & AR & AR$_{50}$ & AR$_{25}$ \\
\drule
ISBNet (fully-sup.)$^\dagger$~\cite{isbnet} & 40.2 & 50.0 & 54.6 & 66.8 & 80.4 & 87.4 \\
Mask3D (fully-sup.)$^\dagger$~\cite{isbnet} & 50.6 & 68.0 & 76.9 & 65.3 & 81.0 & 88.4 \\
\drule
Superpoints$^\dagger$~\cite{superpoints} & 5.0 & 12.7 & 38.9 & - & - & - \\
DBSCAN$^\dagger$~\cite{dbscan} & 1.6 & 5.5 & 32.1 & - & - & -\\
OVIR-3D~\cite{ovir} & 14.4 & 27.5 & 38.8 & - & - & - \\
Mask Clustering~\cite{maskclustering} & 19.2 & 36.6 & 51.7 & - & - & - \\
Open3DIS (2D Only)~\cite{open3dis} & 29.7 & 45.2 & 56.8 & 49.0 & 70.0 & 83.2 \\
\textbf{Ours (2D Only)} & \textbf{33.3} & \textbf{51.9} & \textbf{66.1} & \textbf{50.2} & \textbf{72.2} & \textbf{85.5} \\
\midrule
Open3DIS (2D + 3D)~\cite{open3dis} & 34.6 & 43.1 & 48.5 & 66.2 & 81.6 & 91.4 \\
\textbf{Ours (2D + 3D)} & \textbf{46.6} & \textbf{59.0} & \textbf{64.4} & \textbf{74.0} & \textbf{89.8} & \textbf{95.9} \\
\bottomrule
\end{tabular}}
\end{center}
\vspace{-0.7cm}
\caption{\textbf{Class-agnostic evaluation on the ScanNet200~\cite{scannet}.}
}
\label{tab_ablation_class_agnostic}
\vspace{-0.7cm}
\end{table}

\subsection{Ablation Study}
\vspace{-0.2cm}
\noindent\textbf{Class-agnostic Evaluation.}
To evaluate the quality of generated proposals, we report class-agnostic AP and AR on the ScanNet200 dataset. As shown in Table~\ref{tab_ablation_class_agnostic}, our 2D-only method outperforms all image-based approaches, surpassing the previous SoTA by 3.6\%.
Both our method and OVIR-3D~\cite{ovir} share the intuition of sequentially tracking 3D proposals to progressively grow regions.
However, the performance gap between our method and OVIR-3D is substantial, highlighting the effectiveness of our frame-wise tracklet matching algorithm and iterative merging/removal with refinements.
Furthermore, by leveraging both image-based and point cloud-based 3D proposals, our 2D+3D method achieves the highest ARs across all methods, including fully-supervised approaches.

\begin{table}[h]
\begin{center}
\footnotesize
\vspace{-0.3cm}
\scalebox{0.80}{
\begin{tabular}{l|ccc}
\toprule
Method & AP & AP$_{50}$ & AP$_{25}$ \\
\drule
Tracklet-wise sIOU for Tracking & 34.7 & 54.3 & 69.6 \\
\textbf{Frame-wise sIOU for Tracking} & \textbf{35.1 \textcolor{forestgreen}{(+0.4)}} & \textbf{56.1 \textcolor{forestgreen}{(+1.8)}} & \textbf{70.5 \textcolor{forestgreen}{(+0.9)}} \\
\bottomrule
\end{tabular}}
\vspace{-0.3cm}
\caption{\textbf{Impact of different tracklet matching strategies for aggregation on the subset of the ScanNet200 validation set.} Class-agnostic APs are reported.}
\label{tab:frame_wise}
\end{center}
\vspace{-0.7cm}
\end{table}

\noindent\textbf{Different Tracklet Matching Strategies.}
While we adopt frame-wise sIOU for tracklet matching, some approaches~\cite{ovir} leverage the aggregated 3D mask of each tracklet for matching (i.e., tracklet-wise sIOU for aggregation in Table~\ref{tab:frame_wise}).
Specifically, we maintain aggregated 3D superpoint masks of tracked 2D instances for each tracklet and measure sIOU with new observation by only using co-visible superpoints.
As reported in Table~\ref{tab:frame_wise}, frame-wise sIOU for tracking brings meaningful performance gain for AP$_{50}$ and AP$_{25}$ over the tracklet-wise sIOU while both maintain reasonably good APs.
We conjecture that this is because wrong predictions are always accounted for obtaining aggregated 3D masks in the case of tracklet-wise matching.
However, wrong predictions may not have any impact at all during the frame-wise matching if wrong predictions are distinctive from new observations and other predictions have higher sIOU then them, preventing the wrong predictions from being used for matching.

\begin{table}[ht]
\vspace{-0.2cm}
\begin{center}
\footnotesize
\scalebox{0.95}{
\begin{tabular}{l|ccc}
\toprule
Method & AP & AP$_{50}$ & AP$_{25}$ \\
\drule
Agg. Only & 31.4 & 49.9 & 63.5  \\
+ Iter. Merging/Removal & 31.8 \textcolor{forestgreen}{(+0.4)} & 51.8 \textcolor{forestgreen}{(+1.9)} & 68.5 \textcolor{forestgreen}{(+5.0)} \\
+ Overlap Removal & 33.5 \textcolor{forestgreen}{(+2.1)} & 54.4 \textcolor{forestgreen}{(+4.5)} & 69.8 \textcolor{forestgreen}{(+6.3)} \\
+ Iter. Refine & \textbf{35.1 \textcolor{forestgreen}{(+3.7)}} & \textbf{56.1 \textcolor{forestgreen}{(+6.2)}} & \textbf{70.5 \textcolor{forestgreen}{(+7.0)}} \\
\bottomrule
\end{tabular}}
\vspace{-0.3cm}
\caption{\textbf{Impact of iterative merging/removal, overlap removal, and iterative refinement on the subset of ScanNet200 validation set.} Class-agnostic APs are reported.}
\label{tab:merging}
\end{center}
\vspace{-0.8cm}
\end{table}

\noindent\textbf{Impact of Iterative Merging/Removal}
Unlike existing methods~\cite{open3dis, ovir, sai3d}, our method has an additional false positive suppression step by merging/removing duplicated proposals.
As shown in Table~\ref{tab:merging}, applying iterative merging and removing improves AP$_{25}$ by 5.0\%.
More importantly, if we apply iterative merging with overlap removal in the 2D grounding step, it further brings significant gains in all AP metrics.
This is because overlap removal effectively separates masks spanning multiple instances into each instance or partial masks, which later can be merged/removed.
At last, applying iterative refinement further improves the quality, especially in AP and AP$_{50}$ metrics.

\begin{table}[ht]
\begin{center}
\footnotesize
\vspace{-0.3cm}
\scalebox{0.8}{
\begin{tabular}{l|ccc}
\toprule
Method & mAP & mAP$_{50}$ & mAP$_{25}$\\
\drule
Ours w/ CLIP & 27.5 & 34.7 & 38.2 \\
+ Alpha-CLIP & 30.5 \textcolor{forestgreen}{(+3.0)} & 37.6 \textcolor{forestgreen}{(+2.9)} & 41.1 \textcolor{forestgreen}{(+2.9)} \\
\textbf{+ SMS-based Filtering} & \textbf{32.7 \textcolor{forestgreen}{(+5.2)}} & \textbf{41.4  \textcolor{forestgreen}{(+6.7)}} & \textbf{45.3 \textcolor{forestgreen}{(+7.1)}} \\
\bottomrule
\end{tabular}}
\vspace{-0.3cm}
\caption{\textbf{Impact of Alpha-CLIP and SMS-based filtering in instance classification on the ScanNet200 dataset.}}
\label{tab:inst_classification}
\end{center}
\vspace{-0.8cm}
\end{table}

\noindent\textbf{Impact of Alpha-CLIP and SMS Filtering.}
Table~\ref{tab:inst_classification} demonstrates the effectiveness of using Alpha-CLIP and SMS-based filtering.
As shown, using Alpha-CLIP improves the performance from 27.5 to 30.5 mAP, proving the importance of considering object-centric representation in instance classification.
However, we note that our method with CLIP still surpasses existing baselines by a large margin (i.e., 3.8\% over Open3DIS and 2.8\% over Open-YOLO3D). 
Using SMS-based filtering also brings gains in AP metrics by effectively removing unconfident instances from both image-based and point cloud-based proposals.
Full ablation study on all three datasets can be found in the supplementary materials.

%% file: sec/5_conclusion.tex
\vspace{-0.2cm}
\section{Conclusion}
\vspace{-0.2cm}
\label{sec:conclusion}
In this paper, we carefully combine existing concepts and devise each stage to achieve precise 3D proposal generation and accurate instance classification.
Our robust 3D tracking allows for more precise 3D proposal aggregation.
Also, overlap removal in 2D predictions accompanied with iterative merging/removal enables much fewer false positive 3D proposals, such as overlapped or partial masks.
At last, we adopt Alpha-CLIP to obtain object-centric CLIP representation and remove unconfident 3D proposals by filtering with a standardized maximum similarity score.
Although our method achieves SoTA precision and recall across datasets, our method is computationally intense because heavy 2D foundation models~\cite{alphaclip, grounded_sam, sam} are adopted in our pipeline. Also, we found that our method fails to improve performance on small objects (e.g., ScanNet++ in the supplementary) but rather remain similar to existing approaches. This is because iterative merging and removal is more effective for medium or large objects.
Improving such limitations remains our future work.

%% file: sec/X_suppl.tex
\section{Algorithm on 3D Proposal Merge and Refinement}
We present a more detailed algorithm for merging and refining 3D proposals in Alg.~\ref{alg:merge_incl}. 
This algorithm refines and consolidates a list of tracklets and their associated 3D proposals. It iteratively evaluates the pairwise similarity of proposals using an Intersection-over-Union (IOU) cost matrix and merges those exceeding a defined similarity threshold. During merging, the 3D points and corresponding tracklets are combined, followed by a refinement step that removes low-visibility 3D superpoints from the merged proposal based on their visibility in the tracked 2D masks. A dynamic validity list tracks unmerged proposals and is updated after each merge iteration. The process continues until no proposals meet the merging criteria. This iterative method effectively consolidates overlapping proposals, enhancing the overall accuracy and coherence of 3D proposals.

\begin{algorithm}[!ht]
\centering
\footnotesize
\caption{3D Proposal Merge and Refinement} 
\begin{algorithmic}[1]
\State \textbf{Input:} A list of $K$ tracklets $\{\mathbf{T}_k\}_{k=1}^K$ and associated 3D proposals $\{\mathbf{m}_k\}_{k=1}^{K}$, $k=1,2,\dots, K$
\State \textbf{Output:} A list of filtered tracklets and 3D proposals
\State
\State $K \gets$ \# of 3D proposals 
\State $\mathbf{m} \gets [\mathbf{m}_1, \ldots, \mathbf{m}_K] \in \{0,1\}^{K \times N}$
\State $\mathbf{T} \gets [\mathbf{T}_1, \ldots, \mathbf{T}_K]$
\State $V \gets [1, 2, \dots, K]$ \Comment{Initialize valid proposal indices}
\State $\mathbf{C}^\text{merge} \gets$ getIoUCostMatrix($\mathbf{m}$)
\State $\text{should\_merge} \gets \text{Any}(\mathbf{C}^\text{merge} > \tau^\text{merge})$
\While{$\text{should\_merge} = \text{True}$}
    \State $\text{visited} \gets \text{hashmap}$ \Comment{Track visited proposals}
    \For{row $r = 1, 2, \ldots, K$}
        \If{$\text{visited}[r] = \text{True}$}
            \State \textbf{continue}
        \EndIf
        \For{col $c = 1, 2, \ldots, K$}
            \If{$r = c$ \textbf{or} $\text{visited}[c]$ \textbf{or} $\mathbf{C}^\text{merge}[r, c] \leq \tau^\text{merge}$}
                \State \textbf{continue}
            \EndIf
            \State $\mathbf{m}_r \gets \mathbf{m}_r \cup 
            \mathbf{m}_c$ \Comment{Merge 3D proposals}
            \vspace{+0.5mm}
            \State $\mathbf{T}_r \gets \mathbf{T}_r \cup \mathbf{T}_c$ \Comment{Merge tracklets}
            \State $\mathbf{m}_r \gets$ refine3DProposal($\mathbf{m}_r, \mathbf{T}_r$)\Comment{Refine 3D proposal}
            \State $V \gets  V \setminus \{c\}$ \Comment{Remove merged proposal}
            \State $\text{visited}[c] \gets \text{True}$
        \EndFor
        \State $\text{visited}[r] \gets \text{True}$
    \EndFor
    \State $K \gets$ length($V$) \Comment{Update \# of proposals}
    \State $\mathbf{m} \gets [\mathbf{m}_{i_1}, \ldots, \mathbf{m}_{i_K}], \ i_k \in V$ \Comment{Update 3D proposal list}
    \State $\mathbf{T} \gets [\mathbf{T}_{i_1}, \ldots, \mathbf{T}_{i_K}], \ i_k \in V$ \Comment{Update tracklet list}
    \State $V \gets [1, 2, \dots, K]$ \Comment{Re-initialize valid proposal indices}
    \State $\mathbf{C}^\text{merge} \gets$ getIoUCostMatrix($\mathbf{m}$)
    \State $\text{should\_merge} \gets \text{Any}(\mathbf{C}^\text{merge} > \tau^\text{merge})$
\EndWhile
\end{algorithmic} 
\label{alg:merge_incl}
\end{algorithm}

\begin{figure*}[ht!]
  \centering
  \includegraphics[width=\linewidth]{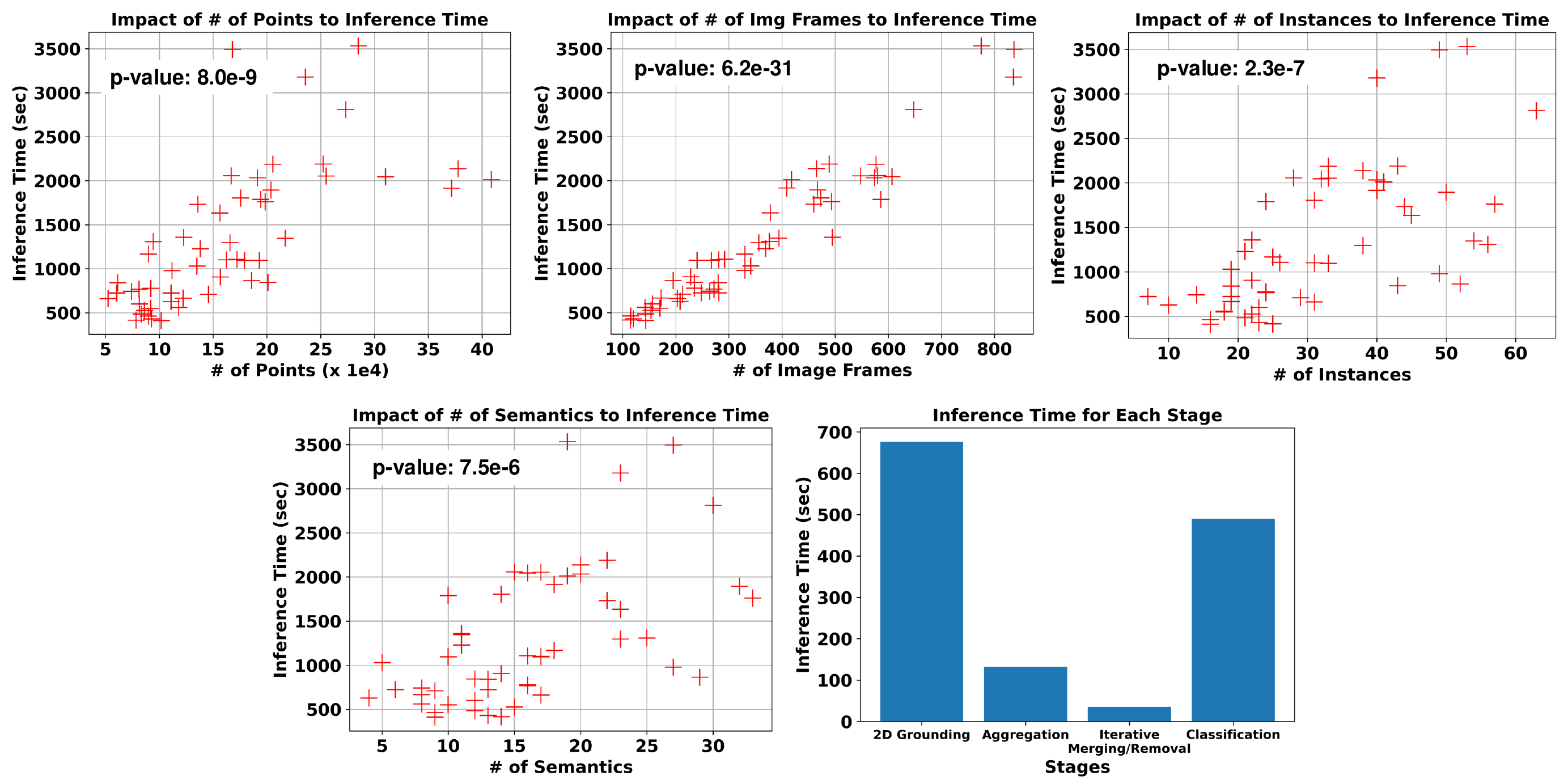}
  \vspace{-0.3cm}
  \caption{\textbf{Computation time analysis of various factors on the subset of ScanNet200 validation set.}}
  \vspace{-0.6cm}
  \label{supp_fig:computation_analysis}
\end{figure*}

\section{More Implementation Details}
We leverage Alpha-CLIP~\cite{alphaclip} with SAM~\cite{sam} for instance classification based on text queries. For the ScanNet200 and Replica datasets, we use instance bounding boxes as queries for mask retrieval from SAM, while for the S3DIS dataset, we utilize subsampled points. This choice is driven by the presence of ``stuff" classes in S3DIS, where subsampled points introduce less noise compared to bounding boxes. For proposal filtering, we adopt different SMS thresholds ($\tau^\text{SMS}$) tailored to each experiment. Performance remains stable within a reasonable range ($\tau^\text{SMS} \in [-1.0, 1.0]$), corresponding to a standard deviation range, as elaborated in Sec.~\ref{sec:supp_exp}.
For instance classification, we select the top 20 visible images for ScanNet200 and Replica and the top 40 images for S3DIS. Following OpenMask3D~\cite{openmask3d}, we use a confidence value of 1.0 to evaluate Average Precision (AP) and Average Recall (AR) metrics. The inference time for our method on the Replica dataset is approximately 597 seconds per scene, closely matching the 547 seconds reported for OpenMask3D~\cite{openmask3d, openyolo3d}. This similarity arises from both methods utilizing SAM and CLIP for instance feature extraction.
In our analysis, instance classification takes up to around 456 seconds, which is 76\% of the computational cost.
All inference times were measured on a single NVIDIA RTX 4090 GPU. Additionally, we employ a text query template, ``\textit{a blurry photo of \{CLASS\_NAME\} in a room}," adapted from CLIP~\cite{clip}.
For Top-K evaluations, we use $K=300$ for 2D-only and 3D-only experiments, and adopt $K=600$ for 2D+3D experiments, following Open3DIS.
OpenYOLO3D adopted $K=600$ for the 3D-only experiment on the ScanNet200 dataset.

\section{Analysis of Computational Cost}
Fig.~\ref{supp_fig:computation_analysis} presents the analysis of the impacts of various factors on the computation time.
We plot graphs to demonstrate the correlation of 1) the number of points in the point cloud, 2) the number of image frames, 3) the number of instances, and 4) the number of different semantic classes present in the scene.
At last, we show the stage-wise computation time of our method.
As shown, we can see meaningful correlations between those factors and computation time.
Also, the 2D grounding step takes the longest computation time in our method, followed by instance classification, 3D proposal aggregation, and iterative merging and removal.

\section{Additional Experiments}
This section presents additional quantitative and qualitative results not included in the main paper.

\subsection{Quantative Results}
\label{sec:supp_exp}
\noindent\textbf{Class-agnostic Evaluation Results on ScanNet++}
We further evaluate our method on the ScanNet++ dataset.
Unlike other datasets, we only evaluate our image-based proposal generation pipeline, excluding point cloud-based 3D proposals.
This is because we experience a non-trivial amount of distributional gap when we apply 3D instance segmentation models trained on other datasets such as ScanNet200.
As reported in Table~\ref{supp_tab:scannet++}, our method demonstrates on-par evaluation results on the AP metric with Open3DIS~\cite{open3dis} + SAM for 2D object grounding.
Under the same VFM (Grounded SAM) for 2D grounding, our method shows better performance in AP$_{50}$ and AP$_{25}$ by 1.7\% and 3.7\%, respectively. 
SAI3D~\cite{sai3d} presents superior results in the AP$_{25}$ metric, surpassing all other methods by a large gap.
We found that our iterative merging/removal step does not contribute to precise 3D proposal generation as much as it used to in other datasets. 
We conjecture that this is because the ScanNet++ dataset includes more small and fine objects that may get removed by merging and removing overlapped and included proposals.
However, our method still maintains reasonable performance, showing on-par results with SoTA methods.

\begin{table}[ht!]
\begin{center}
\footnotesize
\begin{tabular}{cccc}
\toprule
Method & AP & AP$_{50}$ & AP$_{25}$\\
\drule
SAM3D~\cite{sam3d} & 7.2 & 14.2 & 29.4 \\
SAM-guided Graph Cut~\cite{sam_guided} & 12.9 & 25.3 & 43.6 \\
Segment3D~\cite{segment3d} & 12.0 & 22.7 & 37.8 \\
SAI3D~\cite{sai3d} & 17.1 & 31.1 & \textbf{49.5} \\
Open3DIS (G-SAM)$^\dagger$~\cite{open3dis} & 18.2 & 30.7 & 40.6 \\
Open3DIS (SAM)~\cite{open3dis} & \textbf{18.5} & \textbf{33.5} & 44.3 \\
\textbf{Ours (2D Only)} & 18.2 & 32.4 & 44.3 \\
\bottomrule
\end{tabular}
\end{center}
\vspace{-0.7cm}
\caption{\textbf{Class-agnostic evaluation on the ScanNet++ dataset~\cite{scannetpp}.} $^\dagger$numbers are obtained from their official code.}
\label{supp_tab:scannet++}
\end{table}

\begin{table*}[ht!]
\begin{center}
\begin{tabular}{ccccccccc}
\toprule
\multirow{2}{*}{Methods} & \multicolumn{2}{c}{3D Proposals} & \multirow{2}{*}{mAP}& \multirow{2}{*}{mAP$_{50}$} & \multirow{2}{*}{mAP$_{25}$} & \multirow{2}{*}{mAR} & \multirow{2}{*}{mAR$_{50}$} & \multirow{2}{*}{mAR$_{25}$} \\
& Image-based & Point cloud-based & & & & & & \\
\drule
Open3DIS~\cite{open3dis} & \textcolor{forestgreen}{\checkmark} & \textcolor{red}{\xmark} & 17.1 & 27.1 & 36.7 & 24.7 & 37.5 & 49.0 \\
\textbf{Ours (2D Only)} & \textcolor{forestgreen}{\checkmark} & \textcolor{red}{\xmark} & \textbf{22.5} & \textbf{35.0} & \textbf{47.6} & \textbf{32.0} & \textbf{48.7} & \textbf{63.7} \\ 
\midrule
Open3DIS~\cite{open3dis} & \textcolor{red}{\xmark} & \textcolor{forestgreen}{\checkmark} & 24.7 & 30.6 & 35.9 & 34.4 & 41.5 & 47.3 \\
OpenYOLO3D~\cite{openyolo3d} & \textcolor{red}{\xmark} & \textcolor{forestgreen}{\checkmark} & \textbf{37.4} & \textbf{49.4} & \textbf{56.9} & 45.6 & 56.6 & 62.6 \\
\textbf{Ours (3D Only)} & \textcolor{red}{\xmark} & \textcolor{forestgreen}{\checkmark} & \textbf{37.4} & 46.6 & 54.7 & \textbf{47.5} & \textbf{57.2} & \textbf{64.6}\\ 
\midrule
Open3DIS~\cite{open3dis} & \textcolor{forestgreen}{\checkmark} &  \textcolor{forestgreen}{\checkmark} & 27.8 & 33.9 & 39.3 & 44.8 & 53.6 & 60.6 \\
\textbf{Ours (2D + 3D)} & \textcolor{forestgreen}{\checkmark} & \textcolor{forestgreen}{\checkmark} & \textbf{33.5} & \textbf{42.4} & \textbf{47.9} & \textbf{53.6} & \textbf{66.0} & \textbf{72.7}\\ 
\bottomrule
\end{tabular}
\end{center}
\vspace*{-0.6cm}
\caption{\textbf{OV-3DIS results on the S3DIS dataset~\cite{s3dis}.}
The numbers are obtained by using 12 classes, including stuff classes such as floor, ceiling, and wall. Top-1 evaluation protocol is used.}
\label{supp_tab:s3dis}
\end{table*}

\noindent\textbf{S3DIS Results Including ``stuff'' Classes.}
We present results that include "stuff" classes—specifically floor, ceiling, and wall—for evaluation. These classes were excluded from the main paper's evaluation, as our task focuses on segmenting instances, and the notion of instances hardly applies to those classes.
As reported in Table~\ref{supp_tab:s3dis}, our method consistently outperforms baselines in 2D-only and 2D+3D groups. However, in the 3D-only group, our method falls slightly behind OpenYOLO3D in the mAP$_{50}$ and mAP$_{25}$ metrics, primarily due to weaker performance on ``stuff" classes in these metrics.
Nevertheless, our objective is to improve performance on ``thing" classes, which does not necessarily correlate with gains on ``stuff" classes. While this gap could be addressed by incorporating panoptic segmentation methods to handle both types of classes, such exploration is beyond the scope of this work. Importantly, our method achieves SoTA results in all AR metrics across all three groups.

\begin{figure}[t]
  \centering
  \includegraphics[width=0.9\linewidth]{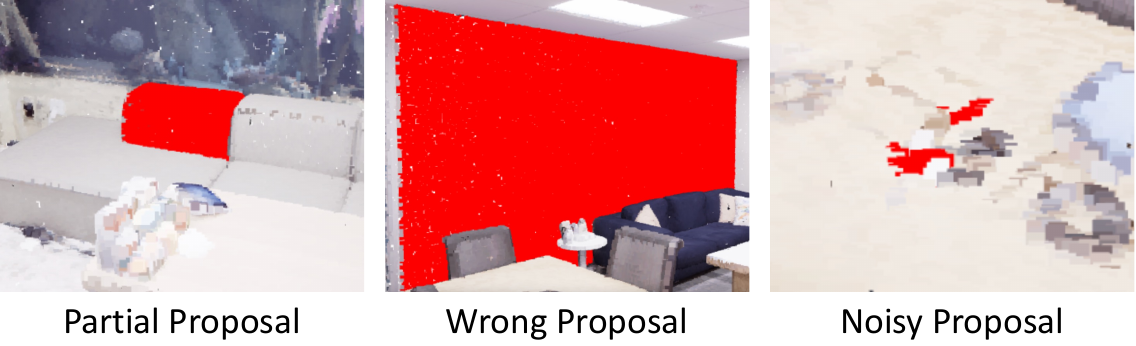}
  \vspace{-0.2cm}
  \caption{\textbf{Visualization of filtered proposals by using the Standardized Maximum Similarity (SMS) score.}
 The SMS score effectively filters out partial proposals (e.g., only part of a sofa is covered), incorrect proposals that do not match any text queries (e.g., "wall" class not included in the evaluation set), and noisy proposals lacking meaningful object representation.
  }
  \vspace{-0.6cm}
  \label{fig:method_sms}
\end{figure}

\begin{figure}[ht]
  \centering
  \includegraphics[width=\linewidth]{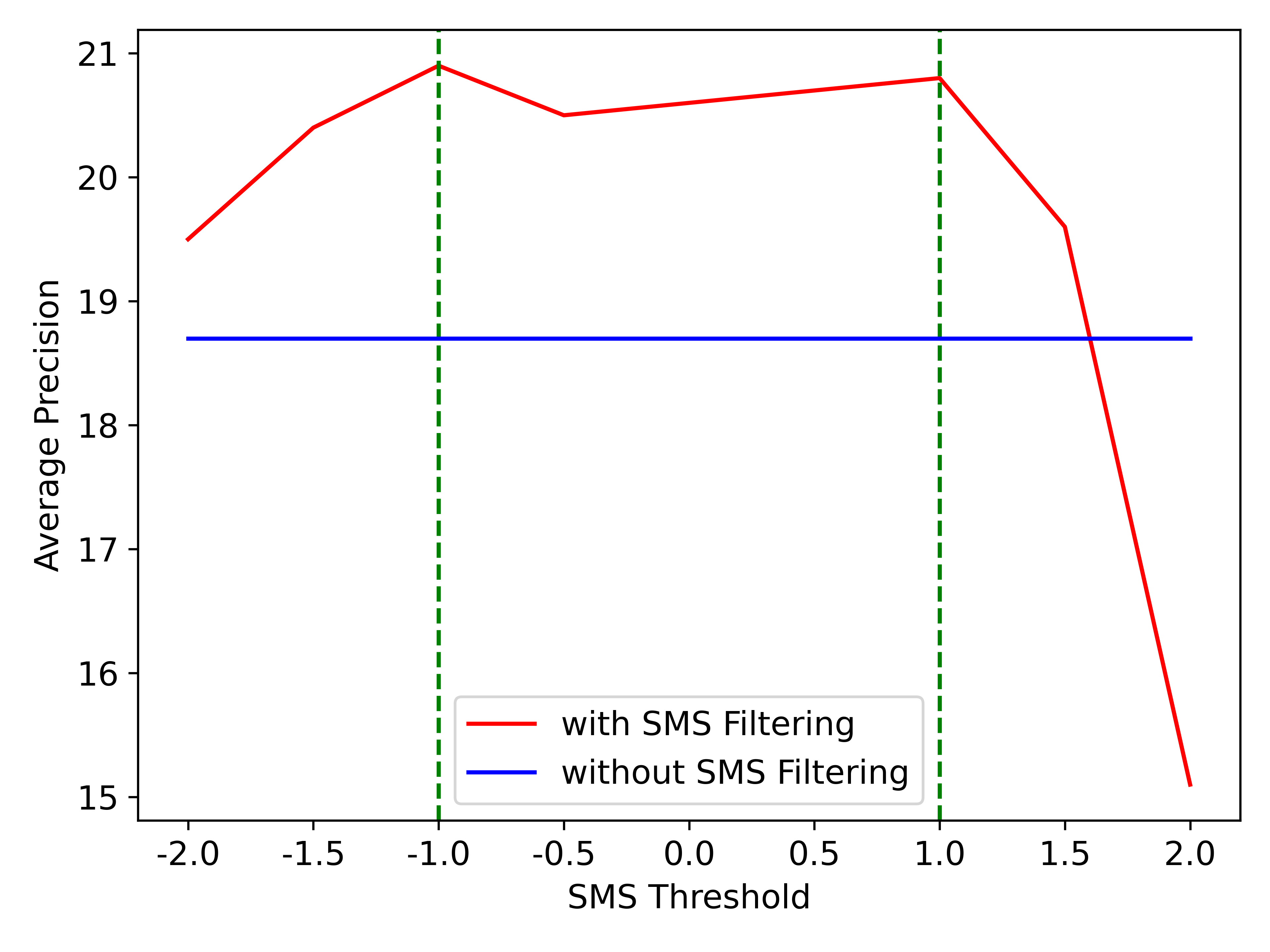}
  \vspace{-0.8cm}
  \caption{\textbf{Impacts of varying SMS filtering thresholds on the AP metric.}
  The red line denotes the AP values across different SMS filtering thresholds, and the blue line indicates AP without using SMS filtering.
  The green vertical lines indicate a desirable range of SMS filtering thresholds.
  The numbers are measured on the Replica dataset.
  }
  \label{fig:supp_sms}
\end{figure}

\noindent\textbf{Ablation Study on SMS filtering.}
Fig.~\ref{fig:supp_sms} illustrates the effect of varying SMS filtering thresholds on the AP metric using the Replica dataset. Within the standard deviation range of [-1, 1], the variance remains minimal compared to the outer ranges, with a maximum gap of only 0.4\%. Notably, applying SMS filtering consistently outperforms the baseline experiment conducted without filtering.

\begin{table}[!ht]
\begin{center}
\footnotesize
\scalebox{0.85}{
\begin{tabular}{cc|cc|cc|cc|cc}
\toprule
$\tau^\text{img}$ & AP & $\tau^\text{inst}$ & AP & $\tau^\text{ref}$ & AP & $\tau^\text{merge}$ & AP & $\tau^\text{incl}$ & AP \\
\drule
0.0 & 26.4 & 0.0 & 28.0 & 0.0 & 33.5 & 0.1 & \textbf{36.1} & 0.5 & \textbf{36.3}  \\
0.1 & \textbf{35.1} & 0.1 & 33.9 & 0.2 & 34.0 & 0.3 & 35.1 & 0.7 & 35.6  \\
0.3 & 27.9 & 0.3 & \textbf{35.1} & 0.4 & \textbf{35.1} & 0.5 & 33.6 & 0.9 & 35.0 \\
0.5 & 16.2 & 0.5 & 29.9 & 0.6 & 33.8 & 0.7 & 33.0 & 0.99 & 35.1  \\
\bottomrule
\end{tabular}
}
\end{center}
\vspace{-0.7cm}
\caption{\textbf{Impact of hyper-parameters on the subset of the ScanNet200 validation set.} Class-agnostic APs are reported.}
\vspace{-0.4cm}
\label{supp_tab:hyper_param}
\end{table}

\noindent\textbf{Impact of Hyper-parameters.}
Table~\ref{supp_tab:hyper_param} demonstrates the impacts of different hyper-parameter values on the generated 3D proposal quality.
As reported, our algorithm is sensitive to $\tau^\text{img}$ and $\tau^\text{inst}$ values because they are applied in the first step of our generation and also provide the basis for later operations.
Nevertheless, our method remains less sensitive to other hyper-parameters.

\noindent\textbf{Alpha CLIP vs CLIP}
Table~\ref{supp_tab:clip} demonstrates impacts of using Alpha-CLIP and SMS filtering for instance classification. In all three datasets, using Alpha-CLIP brings meaningful performance gains consistently across all datasets and metrics.
Applying SMS filtering further improves this, achieving SoTA performance.

\begin{table*}[ht]
\begin{center}
\footnotesize
\begin{tabular}{cl|cccccc}
\toprule
Dataset & Method & mAP & mAP$_{50}$ & mAP$_{25}$ & mAR & mAR$_{50}$ & mAR$_{25}$\\
\drule
\multirow{3}{*}{ScanNet200} & Ours w/ CLIP & 27.5 & 34.7 & 38.2 & 52.4 & 65.6 & 71.8 \\
& + Alpha-CLIP & 30.5 \textcolor{forestgreen}{(+3.0)} & 37.6 \textcolor{forestgreen}{(+2.9)} & 41.1 \textcolor{forestgreen}{(+2.9)} & 57.6 \textcolor{forestgreen}{(+5.2)} & 70.5 \textcolor{forestgreen}{(+4.9)} & 76.5 \textcolor{forestgreen}{(+4.7)} \\
& + SMS Filtering & \textbf{32.7 \textcolor{forestgreen}{(+5.2)}} & \textbf{41.4 \textcolor{forestgreen}{(+6.7)}} & \textbf{45.3 \textcolor{forestgreen}{(+7.1)}} & \textbf{61.4 \textcolor{forestgreen}{(+9.0)}} & \textbf{76.9 \textcolor{forestgreen}{(+11.3)}} & \textbf{83.5 \textcolor{forestgreen}{(+11.7)}} \\
\midrule
\multirow{3}{*}{Replica} & Ours w/ CLIP & 22.4 & 30.0 & 35.6 & 42.8 & 57.1 & 67.7 \\
& + Alpha-CLIP & 25.1 \textcolor{forestgreen}{(+2.7)} & 33.7 \textcolor{forestgreen}{(+3.7)} & 41.7 \textcolor{forestgreen}{(+6.1)} & 47.6 \textcolor{forestgreen}{(+4.8)} & 63.8 \textcolor{forestgreen}{(+6.7)} & 78.0 \textcolor{forestgreen}{(+10.3)}\\
& + SMS Filtering & \textbf{25.7 \textcolor{forestgreen}{(+3.3)}} & \textbf{34.9 \textcolor{forestgreen}{(+4.9)}} & \textbf{42.3 \textcolor{forestgreen}{(+6.7)}} & \textbf{48.8 \textcolor{forestgreen}{(+6.0)}} & \textbf{66.3 \textcolor{forestgreen}{(+9.2)}} & \textbf{79.7 \textcolor{forestgreen}{(+12.0)}}\\
\midrule
\multirow{3}{*}{S3DIS} & Ours w/ CLIP &  29.4 & 39.9 & 45.4 & 45.4 & 60.5 & 67.2 \\
& + Alpha-CLIP & 31.0 \textcolor{forestgreen}{(+1.6)} & 43.1 \textcolor{forestgreen}{(+3.2)} & 49.9 \textcolor{forestgreen}{(+4.5)} & 47.9 \textcolor{forestgreen}{(+2.5)} & 64.7 \textcolor{forestgreen}{(+4.2)} & 72.5 \textcolor{forestgreen}{(+5.3)} \\
& + SMS Filtering & \textbf{31.3 \textcolor{forestgreen}{(+1.9)}} & \textbf{43.5 \textcolor{forestgreen}{(+3.6)}} & \textbf{50.4 \textcolor{forestgreen}{(+5.0)}} & \textbf{48.2 \textcolor{forestgreen}{(+2.8)}} & \textbf{65.1 \textcolor{forestgreen}{(+4.6)}} & \textbf{72.9 \textcolor{forestgreen}{(+5.7)}}\\
\bottomrule
\end{tabular}
\vspace{-0.2cm}
\caption{\textbf{Impact of Alpha-CLIP in instance classification on the ScanNet200, Replica, and S3DIS datasets.}}
\label{supp_tab:clip}
\end{center}
\end{table*}

\begin{figure*}
  \centering
  \includegraphics[width=\linewidth]{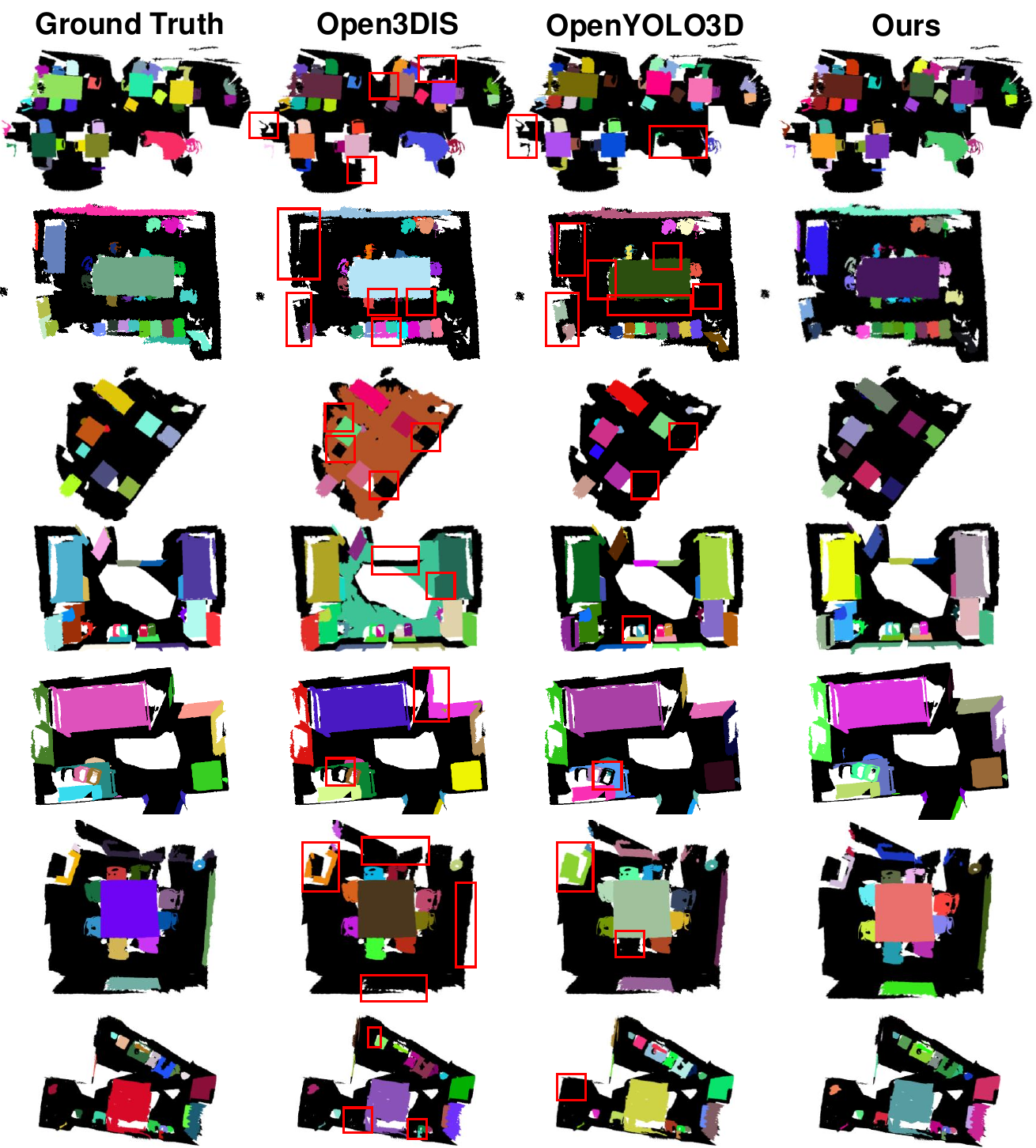}
  \vspace{-0.3cm}
  \caption{\textbf{Exteneded qualitative comparisons on the ScanNet200 dataset.} 
  Black regions indicate empty predictions (\textit{no object}), while red boxes highlight objects missed by other methods but successfully detected by ours. 3D instance masks are colored randomly.
  }
  \vspace*{-0.6cm}
  \label{fig:supp_scannet1}
\end{figure*}

\begin{figure*}
  \centering
  \includegraphics[width=\linewidth]{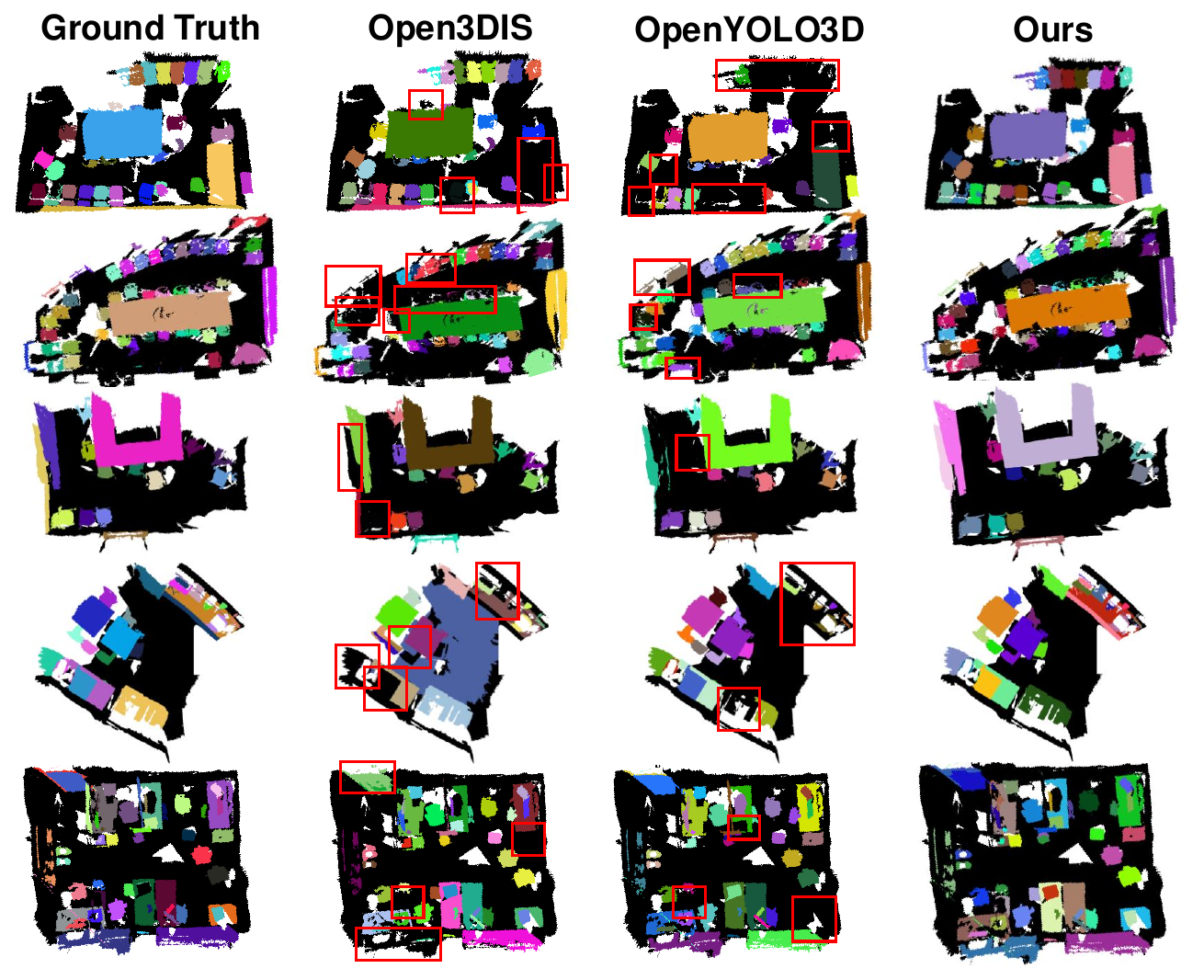}
  \vspace{-0.3cm}
  \caption{\textbf{Exteneded qualitative comparisons on the ScanNet200 dataset.} 
  Black regions indicate empty predictions (\textit{no object}), while red boxes highlight objects missed by other methods but successfully detected by ours. 3D instance masks are colored randomly.
  }
  \vspace*{-0.6cm}
  \label{fig:supp_scannet2}
\end{figure*}

\begin{figure*}
  \centering
  \includegraphics[width=\linewidth]{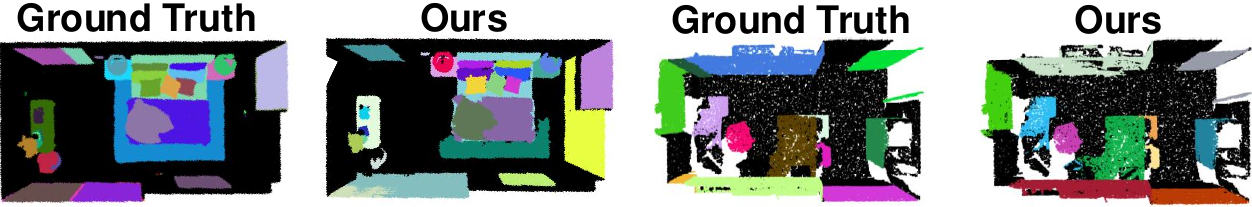}
  \vspace{-0.3cm}
  \caption{\textbf{Qualitative results of our method on the Replica~\cite{replica} (left) and S3DIS~\cite{s3dis} (right) datasets.} 
  Black regions indicate empty predictions (\textit{no object}). 3D instance masks are colored randomly.
  }
  \vspace*{-0.6cm}
  \label{fig:supp_s3dis}
\end{figure*}

\begin{figure*}
  \centering
  \includegraphics[width=\linewidth]{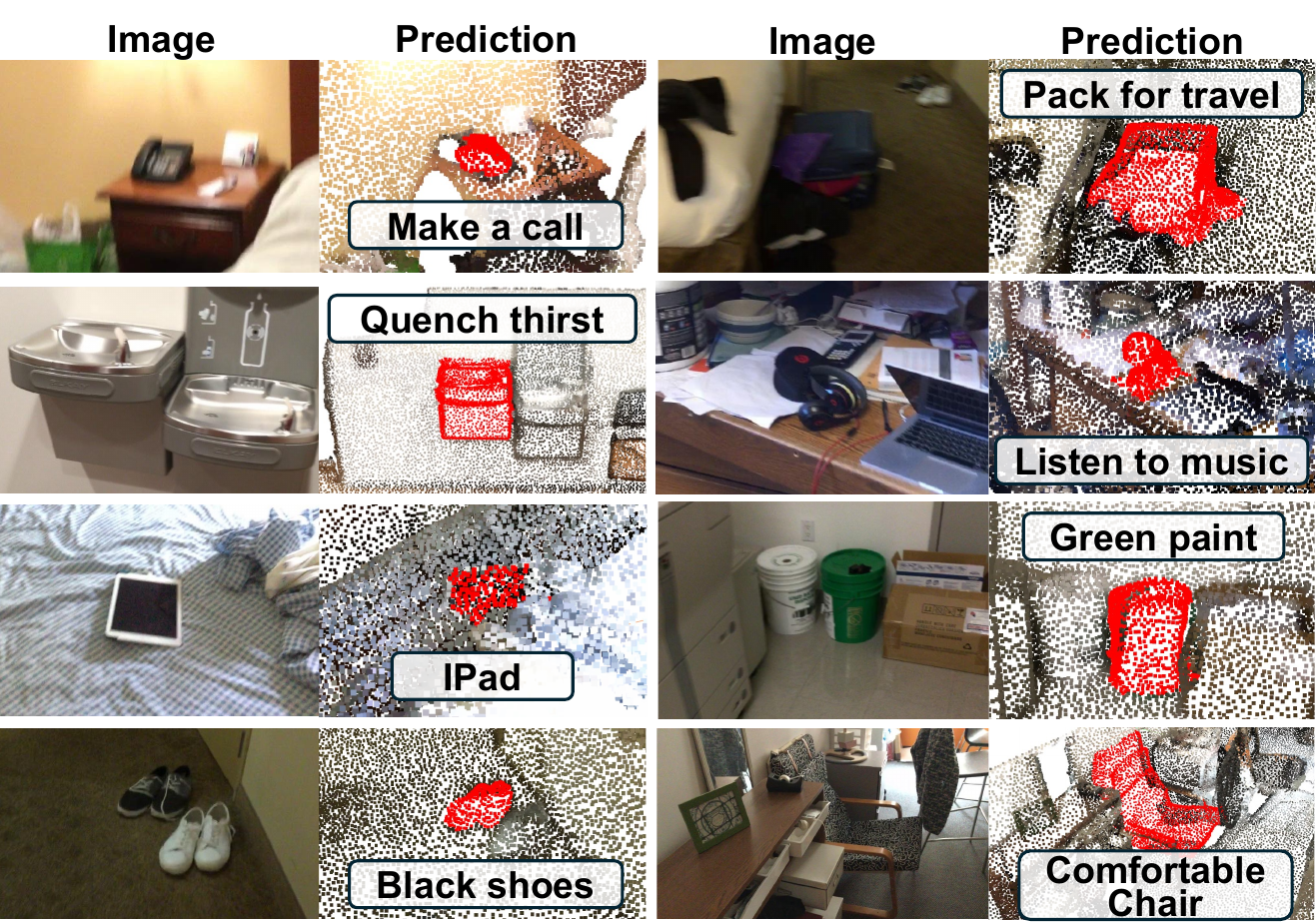}
  \vspace{-0.3cm}
  \caption{\textbf{Exteneded OV-3DIS results with new text queries on the ScanNet200 dataset.} 
  Our method effectively retrieves instances based on functional descriptions and object attributes.
  }
  \vspace*{-0.6cm}
  \label{fig:supp_scannet_query}
\end{figure*}

\subsection{Qualitative Results}
\noindent\textbf{Qualitative Results with Dataset-Provided Text Queries.}
We present more qualitative comparisons on the ScanNet200 dataset~\cite{scannet} in Figs.~\ref{fig:supp_scannet1} and~\ref{fig:supp_scannet2}.
As shown, both Open3DIS and OpenYOLO3D fail to detect certain instances, primarily due to missing image-based 3D proposals or incorrect instance classifications. In contrast, our method not only generates accurate proposals but also classifies them correctly. Open3DIS, in particular, occasionally misidentifies the ``floor" as an instance (see third/fourth and fourth rows of Figs.~\ref{fig:supp_scannet1} and~\ref{fig:supp_scannet2}, respectively), reflecting imperfections in their image-based proposal generation.
We also provide qualitative results on the S3DIS and Replica datasets in Fig.~\ref{fig:supp_s3dis}, demonstrating that our method accurately retrieves most proposals, with only a few instances missed. We attribute these missed instances to either the domain gap between real-world data and the synthetic data from Replica or the training nature of CLIP, which emphasizes foreground regions over background elements such as floors, ceilings, walls, and columns.
Also, in the case of S3DIS, some instances have incomplete masks for large objects, which could be a side effect of our refinements.
This is the limitation of our method, and solving this problem remains our future work.

\noindent\textbf{Qualitative Results with the New Text Queries.}
We visualize more examples of OV-3DIS using new text queries on the ScanNet200 dataset in Fig.~\ref{fig:supp_scannet_query}.
Our method successfully retrieves corresponding instances based on functional descriptions and object attributes such as color, brand name, and other features.